\documentclass{article}

\PassOptionsToPackage{dvipsnames}{xcolor}
\PassOptionsToPackage{round}{natbib}

\usepackage[preprint]{neurips_2021}
\usepackage{multirow}
\usepackage{multicol}
\usepackage{listings}
\usepackage{algpseudocode}
\usepackage{varwidth}
\usepackage{rotating}
\usepackage{subcaption}

\usepackage[utf8]{inputenc} 
\usepackage[T1]{fontenc}    
\usepackage{hyperref}       
\usepackage{url}            
\usepackage{booktabs}       
\usepackage{amsfonts}       
\usepackage{nicefrac}       
\usepackage{microtype}      
\usepackage{xcolor}         
\usepackage{tikz}
\usetikzlibrary{tikzmark}
\usetikzlibrary{matrix, positioning, fit, calc}
\usepackage{graphicx}
\usepackage{booktabs} 
\usepackage{amsmath}
\graphicspath{{./figures/}}

\title{
    Algorithmic Concept-based Explainable Reasoning
}
\author{Dobrik Georgiev\\University of Cambridge\\\texttt{dgg30@cam.ac.uk} \And Pietro Barbiero\\University of Cambridge\\\texttt{pb737@cam.ac.uk} \And Dmitry Kazhdan\\University of Cambridge\\\texttt{dk525@cam.ac.uk} \And Petar Veli\v{c}kovi\'c\\DeepMind\\\texttt{petarv@deepmind.com} \And Pietro Li\`o\\University of Cambridge\\\texttt{pl219@cam.ac.uk}}
\date{}

\newcommand{\CTO}{$\mathcal{C}\rightarrow\mathcal{O}$ }
\newcommand{\CTT}{$\mathcal{C}\rightarrow\mathcal{T}$ }

\newcommand{\Dobotodo}[1]{\textcolor{red}{#1}}

\definecolor{darkolivegreen}{HTML}{556B2F}
\definecolor{mediumslateblue}{HTML}{7B68EE}
\definecolor{Lime}{HTML}{00FF00}

\begin{document}

\maketitle


\begin{abstract}
Recent research on graph neural network (GNN) models successfully applied GNNs
to classical graph algorithms and combinatorial optimisation problems. This has
numerous benefits, such as allowing applications of algorithms when
preconditions are not satisfied, or reusing learned models when sufficient
training data is not available or can't be generated. Unfortunately, a key
hindrance of these approaches is their lack of explainability, since GNNs are
black-box models that cannot be interpreted directly. In this work, we address
this limitation by applying existing work on concept-based explanations to GNN
models. We introduce \textit{concept-bottleneck GNNs}, which rely on
a modification to the GNN readout mechanism. Using three case
studies we demonstrate that: (i) our proposed model is capable of accurately
learning concepts and extracting propositional formulas based on the learned
concepts for each target class; (ii) our concept-based GNN models achieve
comparative performance with state-of-the-art models; (iii) we can derive
global graph concepts, without explicitly providing any supervision on
graph-level concepts. 
\end{abstract}

\section{Introduction}

Graph neural networks (GNNs) have successfully been applied to problems involving data
with irregular structure, such as quantum chemistry
\citep{Gilmer2017Neural}, drug discovery \citep{Stokes2020ADeep}, social
networks \citep{Pal2020PinnerSage} and physics simulations
\citep{Battaglia2016Interaction}. One of the latest areas of GNN research
focuses on using GNNs for emulation of classical algorithms
\citep{Cappart2021Combinatorial}. 
In particular, this research explored applications of GNNs to iterative algorithms
\citep{Velickovic2020Neural, Georgiev2020Neural}, pointer-based data structures
\citep{Velickovic2020Pointer, Strathmann2021Persistent}, and even planning
tasks \citep{Deac2020XLVIN}. Importantly, these works demonstrate that GNNs are capable of \emph{strongly generalising} to input graphs much larger than the ones seen during training. 



Unfortunately, in all of the aforementioned cases, these state-of-the-art GNN
models are black-boxes, whose behaviour cannot be understood/intepreted
directly. In practice, this can lead to a lack of trust in such models, making
it challenging to apply and regulate these models in safety-critical
applications, such as healthcare.  Furthermore, this lack of interpretability
also makes it difficult to extract the knowledge learned by such models, which
prevents users from better understanding the corresponding tasks \citep{adadi2018peeking,
molnar2020interpretable, doshi2017towards}.




Recent work on Explainable AI (XAI) introduced a novel
type of Convolutional Neural Network (CNN) explanation approach, referred to as
\textit{concept-based explainability} \citep{Koh2020Concept, kazhdan2020now, ghorbani2019towards, kazhdan2021disentanglement}. Concept-based explanation approaches provide model
explanations in terms of human-understandable units, rather than individual
features, pixels, or characters (e.g., the concepts of a wheel and a door are
important for the detection of cars) \citep{kazhdan2020now}. In particular, work on Concept Bottleneck
Models (CBMs) relies on concepts and introduces a novel type of
interpretable-by-design CNN, which perform input processing in two distinct
steps: computing a set of concepts from an input, and then computing the output
label from the concepts \citep{Koh2020Concept}.

\begin{figure}[t]
    \centering
    \begin{tabular}{p{0.28\textwidth} | p{0.28\textwidth}| p{0.31\textwidth}} 
        \multicolumn{1}{c|}{BFS}
&
\multicolumn{1}{c|}{Parallel coloring}
&
\multicolumn{1}{c}{Kruskal's algorithm}
\\
\hline
{\usetikzlibrary{arrows, shapes,  calc, fit, decorations.pathmorphing, positioning}
\tikzstyle{vertex}=[circle,draw, minimum size=20pt,inner sep=0pt]
\tikzstyle{vis vertex} = [vertex, fill=mygreen!60, minimum size=20pt,inner sep=0pt]
\definecolor{mygreen}{rgb}{0.00784313725490196, 0.6196078431372549, 0.45098039215686275}

\begin{tikzpicture}[scale=0.60, auto,swap, every node/.style={scale=0.70}]
\tikzset{VertexStyle/.style =
    {draw, shape=circle,minimum size=25pt,inner sep=0pt, scale=0.68}
}
\tikzset{LabelStyle/.style = {text=black, font=\small, below, sloped}}
\tikzset{NSLabelStyle/.style = {text=black, font=\small, align=center}}
\tikzset{MinDistStyle/.style = {transform canvas={xshift=1.5mm, yshift=1.5mm}, text=mygreen}}
\node[mygreen] (pietro) at (-1.5, 3) {visited node};
\node (fake) at (-2.0, 0) {};
\node[vertex] (G-2) at (0, 0) {e};
\node[vertex, below left=9pt and 3pt of G-2] (G-3) {d};
\node[vertex, below right=9pt and 5pt of G-2] (G-4) {c};
\node[vis vertex, right=8pt of G-4] (G-5) {b};
\node[vis vertex, above=12pt of G-5] (G-6) {a};
\node[vis vertex, above left=12pt and 12pt of G-6] (G-7) {f};
\tikzset{EdgeStyle/.style = {thick, }}
\draw[EdgeStyle] (G-3) to (G-4);
\draw[EdgeStyle] (G-2) to (G-4);
\draw[EdgeStyle] (G-2) to (G-3);
\draw[EdgeStyle] (G-2) to (G-7);
\draw[EdgeStyle] (G-4) to (G-5);
\draw[EdgeStyle] (G-5) to (G-6);
\draw[EdgeStyle] (G-7) to (G-6);
\draw[EdgeStyle] (G-7) to (G-4);
\node[draw,inner sep=4pt,fit=(G-4), color=RedViolet] (fitreal) {};

\node[below=20pt of G-4, align=center] (rc) {has NOT been vis.\\has visited neighbour};
\node[below=20pt of rc, align=center, scale=1.4] (ro) {$v_c=1$};

\node (fr) at ($(fitreal.south)-(0.00, -0.05)$) {};
\draw[EdgeStyle, -stealth, densely dotted, RedViolet] (fr) to (rc);

\draw[EdgeStyle, -stealth, densely dotted, RedViolet] (rc) to (ro);
\node[rectangle, fill=black!25, below=4pt of G-4, align=center] (CBGNN) {CBGNN};
\node[rectangle, fill=black!25, below=3pt of rc, align=center] (CBGNN) {CBGNN};
\end{tikzpicture}} 
& {\usetikzlibrary{arrows, shapes, shapes.geometric,  calc, fit, decorations.pathmorphing, positioning}
\tikzstyle{vertex}=[circle,draw, minimum size=20pt,inner sep=0pt]
\tikzstyle{vis vertex} = [vertex, fill=mygreen!24, minimum size=20pt,inner sep=0pt]
\definecolor{mygreen}{rgb}{0,0.6,0}
\definecolor{sc1}{rgb}{0.00392156862745098,0.45098039215686275,0.6980392156862745}
\definecolor{sc2}{rgb}{0.8705882352941177, 0.5607843137254902, 0.0196078431372549}
\definecolor{sc3}{rgb}{0.00784313725490196, 0.6196078431372549, 0.45098039215686275}
\definecolor{sc4}{rgb}{0.8352941176470589, 0.3686274509803922, 0.0}
\definecolor{sc5}{rgb}{0.8, 0.47058823529411764, 0.7372549019607844}

\begin{tikzpicture}[scale=0.60, auto,swap, every node/.style={scale=0.70}]
\tikzset{VertexStyle/.style =
    {draw, shape=circle,minimum size=25pt,inner sep=0pt, scale=0.68}
}
\tikzset{LabelStyle/.style = {text=black, font=\small, below, sloped}}
\tikzset{NSLabelStyle/.style = {text=black, font=\small, align=center}}
\tikzset{MinDistStyle/.style = {transform canvas={xshift=1.5mm, yshift=1.5mm}, text=mygreen}}
\node (fake) at (-3.0, 0) {};
\node[vertex] (G-1) at (0, 0) {33};
\node[vertex, above left=18pt and 5pt of G-1] (G-2) {2};
\node[vertex, fill=sc3!60, above right=25pt and 3pt of G-1] (G-3) {100};
\node[vertex, fill=sc1!60, above left=5pt and 19pt of G-1] (G-4) {44};
\node[vertex, fill=sc2!60, below left=2pt and 18pt of G-1] (G-5) {58};
\node[vertex, fill=sc1!60, below right=2pt and 18pt of G-1] (G-6) {248};
\node[vertex, fill=sc2!60, above right=-2pt and 15pt of G-1] (G-7) {155};
\tikzset{EdgeStyle/.style = {thick, }}
\draw[EdgeStyle] (G-1) to (G-2);
\draw[EdgeStyle] (G-1) to (G-3);
\draw[EdgeStyle] (G-1) to (G-4);
\draw[EdgeStyle] (G-1) to (G-5);
\draw[EdgeStyle] (G-1) to (G-6);
\draw[EdgeStyle] (G-1) to (G-7);
\draw[EdgeStyle] (G-2) to (G-3);
\draw[EdgeStyle] (G-2) to (G-4);
\draw[EdgeStyle] (G-4) to (G-5);
\draw[EdgeStyle] (G-5) to (G-6);
\draw[EdgeStyle] (G-6) to (G-7);
\draw[EdgeStyle] (G-7) to (G-3);
\draw[EdgeStyle, bend right=90] (G-6) to (G-3);

\node[scale=1.2, above left=45pt and 5pt of G-1] (available colors){Colors:};
\node[rectangle, scale=1.3, draw, fill=sc1!60, right=5pt of available colors] (c1) {};
\node[rectangle, scale=1.3, draw, fill=sc2!60, right=4pt of c1] (c2) {};
\node[rectangle, scale=1.3, draw, fill=sc3!60, right=4pt of c2] (c3) {};
\node[rectangle, scale=1.3, draw, fill=sc4!60, right=4pt of c3] (c4) {};
\node[rectangle, scale=1.3, draw, fill=sc5!60, right=4pt of c4] (c5) {};
\node[left=0.01mm of G-1] (n) {$n$};
\node[above=1pt of c1] (i1) {1};
\node[above=1pt of c2] (i2) {2};
\node[above=1pt of c3] (i3) {3};
\node[above=1pt of c4] (i4) {4};
\node[above=1pt of c5] (i5) {5};
\node[draw,inner sep=4pt,fit=(G-1), color=RedViolet] (fitreal) {};
\node (fr) at ($(fitreal.south)-(0.00, -0.05)$) {};
\node[below=22pt of G-1, align=center] (rc) {NOT colored, has priority\\color 1, 2 and 3 seen\\color 4 NOT seen};
\node[below=18pt of rc, align=center, scale=1.4] (ro) {$c_n=4$};
\draw[EdgeStyle, -stealth, densely dotted, RedViolet] (fr) to (rc);
\draw[EdgeStyle, -stealth, densely dotted, RedViolet] (rc) to (ro);
\node[rectangle, fill=black!25, above=5pt of rc, align=center] (CBGNN) {CBGNN};
\node[rectangle, fill=black!25, above=6pt of ro, align=center] (CBGNN) {CBGNN};
\end{tikzpicture}} & {\usetikzlibrary{arrows, shapes, shapes.geometric,  calc, fit, decorations.pathmorphing, positioning}
\tikzstyle{vertex}=[circle,draw, minimum size=20pt,inner sep=0pt]
\tikzstyle{vis vertex} = [vertex, fill=mygreen!24, minimum size=20pt,inner sep=0pt]
\definecolor{mygreen}{rgb}{0,0.6,0}
\definecolor{sc1}{rgb}{0.00392156862745098,0.45098039215686275,0.6980392156862745}
\definecolor{sc2}{rgb}{0.8705882352941177, 0.5607843137254902, 0.0196078431372549}
\definecolor{sc3}{rgb}{0.00784313725490196, 0.6196078431372549, 0.45098039215686275}
\definecolor{sc4}{rgb}{0.8352941176470589, 0.3686274509803922, 0.0}
\definecolor{sc5}{rgb}{0.8, 0.47058823529411764, 0.7372549019607844}

\begin{tikzpicture}[scale=0.60, auto,swap, every node/.style={scale=0.70}]
\tikzset{VertexStyle/.style =
    {draw, shape=circle,minimum size=25pt,inner sep=0pt, scale=0.68}
}
\tikzset{LabelStyle/.style = {text=black, font=\small, below, sloped}}
\tikzset{NSLabelStyle/.style = {text=black, font=\small, align=center}}
\tikzset{MinDistStyle/.style = {transform canvas={xshift=1.5mm, yshift=1.5mm}, text=mygreen}}
\node[vertex] (G-1) at (0, 0) {a};
\node[vertex, right=25pt of G-1] (G-2) {b};
\node[vertex, below=25pt of G-1] (G-3) {c};
\node[above left=0pt and 8pt of G-3] (g3) {};
\node[vertex, below right=29pt and 29pt of G-1] (G-4) {d};
\node[above right=0pt and 8pt of G-4] (g4) {};
\node[vertex, above right=15pt and 10pt of G-1] (G-5) {e};
\node[sc3, above left=0pt and 10pt of G-5] (sele) {edge in MST};
\node[sc5!70, below=-1pt of sele] (unsele) {visited edge};
\node[below right=-10pt and 30pt of sele, align=center] (sets) {$S_1=\{a, b, e, c\}$ \\ $S_2=\{d\}$};
\tikzset{EdgeStyle/.style = {thick, }}
\draw[EdgeStyle, sc3] (G-1) to node[above] {4} (G-2);
\draw[EdgeStyle, sc3] (G-1) to node[left] {2} (G-3);
\draw[EdgeStyle] (G-1) to node[below] {8} (G-4);
\draw[EdgeStyle, sc5!70] (G-1) to node[left] {5} (G-5);
\draw[EdgeStyle] (G-2) to node[right] {12} (G-4);
\draw[EdgeStyle, sc3] (G-2) to node[right] {1} (G-5);
\draw[EdgeStyle] (G-3) to node[below] (hui) {6} (G-4);
\node[draw,inner sep=2pt,fit=(g3) (hui) (g4), color=RedViolet] (fitreal) {};
\node[below=20pt of fitreal, align=center] (rc) {lighter edges visited\\nodes in different sets};
\node[below=20pt of rc, align=center, scale=1.4] (ro) {$m_{cd}=1$};
\node (fr) at ($(fitreal.south)-(0.00, -0.05)$) {};
\draw[EdgeStyle, -stealth, densely dotted, RedViolet] (fr) to (rc);
\draw[EdgeStyle, -stealth, densely dotted, RedViolet] (rc) to (ro);
\node[rectangle, fill=black!25, below=4pt of fitreal, align=center] (CBGNN) {CBGNN};
\node[rectangle, fill=black!25, below=3pt of rc, align=center] (CBGNN) {CBGNN};
\end{tikzpicture}} \\
    
    \end{tabular}
    \caption{
        An overview of our Concept Bottleneck Graph Neural Network (CBGNN)
        approach. Importantly, CBGNN models can be trained to extract concept
        information for a given task as well as algorithm rules. We give
        examples of 3 algorithms, showing how CBGNN extract concepts from
        the input data and then uses these to compute the output.
    }\label{fig:visual_abstract}
\end{figure}
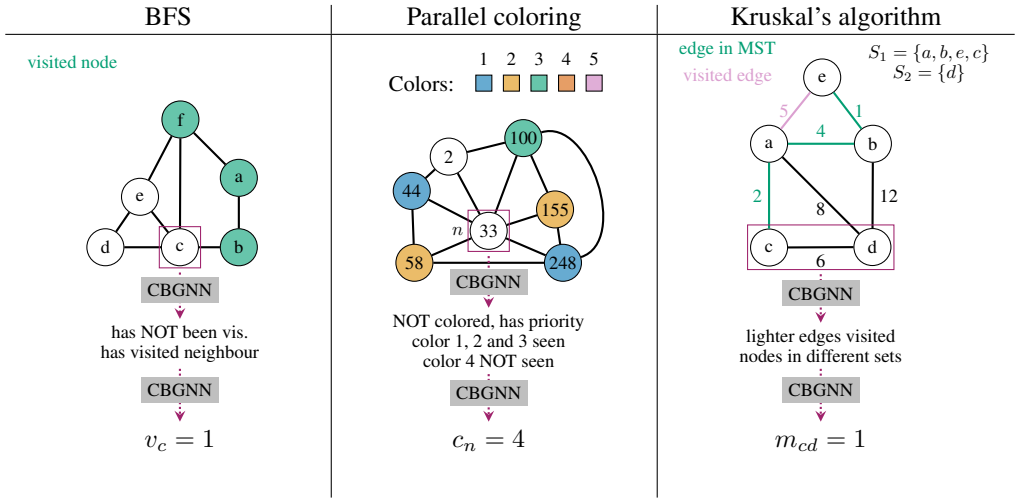

In this paper, we apply the idea of CBMs to GNN models, by introducing Concept
Bottleneck Graph Neural Networks (CBGNNs). In particular, we rely on the
\textit{encode-process-decode} paradigm \citep{Hamrick2018Relational}, and
apply concept bottleneck layers before the output of GNN models -- see Figure
\ref{fig:visual_abstract}. By doing this we are able to extract
update/termination rules for the step updates of \emph{step-level}
combinatorial optimisation approaches \citep{Velickovic2020Neural,
Velickovic2020Pointer, Deac2020XLVIN, Strathmann2021Persistent}.




Importantly, we show that by relying on a suitable set of concepts and
supervising on them, we are capable of deriving the rules of classical
algorithms such as breadth-first search \citep{moore1959shortest}, and
Kruskal's algorithm \citep{Kruskal1956Shortest}, as well as more advanced
heuristics such as parallel graph coloring \citep{Jones1993APG}.  Furthermore,
we present an approach to utilise node-level concepts for extracting
graph-level rules. Our evaluation experiments demonstrate that all of our
extracted rules strongly generalise to graphs of $5\times$ larger size.

To summarise, we make the following contributions in this work: 

\begin{itemize}
    \item {Present \textit{Concept Bottleneck Graph Neural Networks} (CBGNN), a novel type of GNN relying on intermediate concept processing. To the best of our knowledge, this is the first work to apply concept bottleneck approaches to GNNs.}
    
    \item{Quantitatively evaluate our approach using three different case-studies (BFS, graph colouring, and Kruskal's), showing that our CBGNN approch is capable of achieving performance on-par with that of existing state-of-the-art}
    
    \item{Qualitatively evaluate our approach, by demonstrating how the concepts utilised by CBGNN models can be used for providing rules summarising the heuristics the CBGNN has learned}
    
\end{itemize}

\section{Related work}

\textbf{GNN Explainability} Recent work began exploring applications of
XAI techniques in the context of GNNs. For instance, work in
\cite{pope2019explainability, baldassarre2019explainability, schnake2020higher}
adapt feature-importance gradient-based approaches used for CNN applications
(such as Class Activation Mappings, or Layer-wise Relevance Propagation) to
GNNs, in order to identify the most important nodes/subgraphs responsible for
individual predictions. Alternatively, works in \cite{ying2019gnnexplainer,
vu2020pgm, luo2020parameterized} focus on more complex approaches unique to GNN
explainability, such as those based on mutual information maximisation, or
Markov blanket conditional probabilities of feature explanations. Importantly,
these works focus on GNN tasks and benchmarks involving social networks,
chemistry, or drug discovery, instead of focusing on combinatorial optimisation
tasks, which is the focus of this work. Furthermore, these works focus on
explaining pre-trained GNNs in a post-hoc fashion, whereas we focus on building
GNN models interpretable-by-design. Finally, these works focus on
feature-importance-based explanation approaches (i.e. returing relative
importance of input nodes/subgraphs), whereas we rely on concept-based
explanation approaches instead.

\textbf{Concept-based Explainability} A range of existing works have explored
various concept-based explanations applied to CNN models. For instance, work in
\cite{ghorbani2019towards, kazhdan2020now, yeh2019completeness} introduce
approaches for extracting concepts from pre-trained CNNs in an unsupervised, or
semi-supervised fashion. Work in \cite{chen2020concept, Koh2020Concept} rely on
concepts for introducing CNN models interpretable-by-design, performing
processing in two distinct steps: concept extraction, and label prediction.
Other works on concepts include studying the connection between concepts and
disentanglement learning \citep{kazhdan2021disentanglement}, as well as using
concepts for data distribution shifts \citep{wijaya2021failing}. Importantly,
these works explore concepts exclusively in the context of CNNs, with
\cite{kazhdan2020meme} being the only work exploring concepts in the context of
RNN models. In this work, we focus on concept-based explainability for GNNs,
where, similar to \citet{Koh2020Concept}, the concepts are human-specified.

\textbf{Combinatorial Optimisation for GNNs} 
Following the hierarchy defined in \citet{Cappart2021Combinatorial}, our work
classifies as a step-level approach. We  directly extend on
\citet{Velickovic2020Pointer, Velickovic2020Neural}, therefore we use the
models presented in these works as baselines. We \emph{do not} compare
our model to an algorithm-level combinatorial optimisation approaches
\citep{Xu2020What, Tang2020Towards, Joshi2020Learning} or unit-level ones
\citep{Yan2020Neural} for the following reasons: \emph{Algorithm-level
approaches usually give one output per data sample} (rather than one output per
step), but rules/invariants of a given algorithm come from how the iteration
proceeds making algorithm-level combinatorial optimisation less suitable for
a concept bottleneck. \emph{Unit-level learning focuses on learning primitive
units of computation}, such as taking maximum or merging lists and then
combining these manually -- having explanations at this level would not be of
great benefit. To the best of our knowledge, only \citet{Velickovic2020Pointer} attempted to
explain GNN predictions, using GNNExplainer \citep{ying2019gnnexplainer}.
However, their model (i) was not explainable by design and (ii) required further
optimisation \emph{for a single sample} to give a \emph{local explanation}.
All other previous works operated in a black-box fashion and did not consider
explainability of the learnt models.

\section{Methodology}


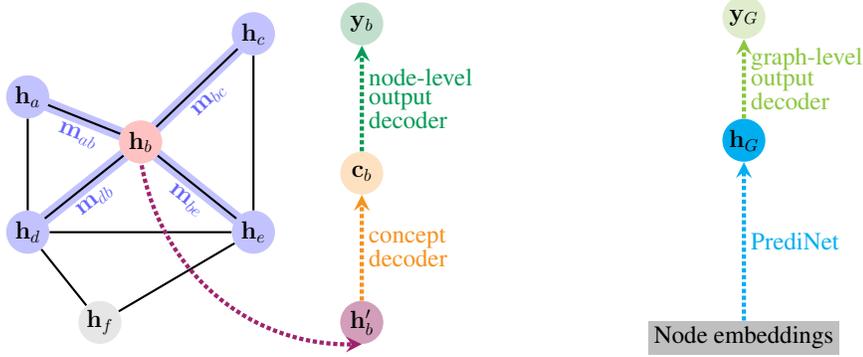
\begin{figure}[t]
    \centering
    \begin{subfigure}{0.49\textwidth}
        \centering
        \usetikzlibrary{arrows,shapes, positioning, calc}

\definecolor{mygreen}{rgb}{0,0.6,0}

\pgfdeclarelayer{background}
\pgfsetlayers{background,main}

\tikzstyle{vertex}=[circle,fill=black!10,minimum size=25pt,inner sep=0pt]
\tikzstyle{red vertex} = [vertex, fill=red!24]
\tikzstyle{blue vertex} = [vertex, fill=blue!24]
\tikzstyle{output vertex} = [vertex, fill=ForestGreen!24]
\tikzstyle{updated vertex} = [vertex, fill=RedViolet!40]
\tikzstyle{conceptualised vertex} = [vertex, fill=BurntOrange!24]
\tikzstyle{edge} = [draw,thick,-]
\tikzstyle{selected edge} = [draw,line width=5pt,-,blue!24]

\begin{tikzpicture}[scale=1.20, auto,swap, every node/.style={scale=0.65}]
    \foreach \pos/\name in {{(-0.5,1.5)/a}, {(0.75,1)/b}, {(2.,2.2)/c},
                            {(-0.5,0)/d}, {(2.,0)/e}, {(0.3,-1)/f}}
        \node[vertex] (\name) at \pos {};
        
    \foreach \source/ \dest /\weight in {b/a/7, c/b/8,d/a/5,d/b/9,
                                         e/b/7, e/c/5,e/d/15,
                                         f/d/6,f/e/8
                                         }
        \path[edge] (\source) -- (\dest);
         
    \foreach \vertex / \fr in {b/4}
        \path node[red vertex] at (\vertex) {\Large $\mathbf{h}_b$};
    \foreach \vertex / \fr in {a/4, c/4, d/4, e/5}
        \path node[blue vertex] at (\vertex) {\Large $\mathbf{h}_{\vertex}$};
    \foreach \vertex in {f}
        \path node at (\vertex) {\Large $\mathbf{h}_{\vertex}$};

    \begin{pgfonlayer}{background}
        \foreach \source / \dest in {b/c,d/b,a/b,b/e}
        \path[selected edge, sloped] (\source.center) edge node [blue!60] {\Large $\mathbf{m}_{\source\dest}$} (\dest.center);
    \end{pgfonlayer}

    
    \node[vertex] (b1) at (3.2,-1) {};
    \foreach \vertex / \fr in {b1/4}
        \path node[updated vertex] at (\vertex) {\Large $\mathbf{h}'_b$};
        
    \node (hui) at ($(b.south)-(0.00, -0.05)$) {};
    \draw[-stealth, densely dotted, ultra thick, RedViolet, sloped] (hui) edge[bend right=40] node {} (b1.south);

    \node[vertex, above = 40pt of b1, conceptualised vertex] (b2) {\Large $\mathbf{c}_b$};

    \draw[-stealth, densely dotted, ultra thick, BurntOrange] (b1) edge node[align=left] {\Large concept\\\Large decoder} (b2);

    \node[vertex, above = 40pt of b2, output vertex] (b3) {\Large $\mathbf{y}_b$};

    \draw[-stealth, densely dotted, ultra thick, ForestGreen] (b2) edge node[align=left] {\Large node-level\\\Large output\\\Large decoder} (b3);
\end{tikzpicture}
    \end{subfigure}
    \begin{subfigure}{0.48\textwidth}
        \centering
        \usetikzlibrary{arrows,shapes, positioning, decorations.pathmorphing, shapes.geometric}

\definecolor{mygreen}{rgb}{0,0.6,0}

\pgfdeclarelayer{background}
\pgfsetlayers{background,main}

\tikzstyle{vertex}=[circle,fill=black!25,minimum size=25pt,inner sep=0pt]
\tikzstyle{red vertex} = [vertex, fill=red!24]
\tikzstyle{blue vertex} = [vertex, fill=blue!24]
\tikzstyle{output vertex} = [vertex, fill=LimeGreen!24]
\tikzstyle{updated vertex} = [vertex, fill=RedViolet!40]
\tikzstyle{conceptualised vertex} = [vertex, fill=Aquamarine]
\tikzstyle{edge} = [draw,thick,-]
\tikzstyle{selected edge} = [draw,line width=5pt,-,blue!24]

\begin{tikzpicture}[scale=1.20, auto,swap, every node/.style={scale=0.65}]
    \node[rectangle, fill=black!25] (node emb) at (2, -0.5) {\Large Node embeddings};

    \node[circle, fill=Cyan, minimum size=25pt,inner sep=0pt, above=60pt of node emb] (hG) {};
         
    \path node at (hG) {\Large $\mathbf{h}_G$};

    \node[output vertex, above=30pt of hG] (oG) {\Large $\mathbf{y}_G$};
    \draw[-stealth, densely dotted, ultra thick, Cyan] (node emb) edge node[align=left] {\Large PrediNet} (hG);
    \draw[-stealth, densely dotted, ultra thick, LimeGreen] (hG) edge node[align=left] {\Large graph-level\\\Large output\\\Large decoder} (oG);

\end{tikzpicture}
    \end{subfigure}
    \caption{
        A high level overview of our GNN architecture: \textbf{Left:} (Eqns. \ref{eq:1}-\ref{eq:4}) To produce
        node-level outputs, messages from neighbouring nodes
        (\textcolor{blue!60}{$\mathbf{m}_{ij}$}) are combined with current node
        representation (\Dobotodo{$\mathbf{h}_b$}), resulting in an updated
        representation (\textcolor{RedViolet}{$\mathbf{h}'_b$}). Concepts
        (\textcolor{BurntOrange}{$\mathbf{c}_b$}) are then extracted from the
        updated representations, and node-level outputs
        (\textcolor{ForestGreen}{$\mathbf{y}_b$}) are extracted from the
        concepts.
        \textbf{Right:} (Eqns. \ref{eq:5}-\ref{eq:8}) A graph-level embedding
        (\textcolor{Cyan!100}{$\mathbf{h}_G$}) is obtained by passing the node embeddings
        through PrediNet. We extract graph-level outputs
        \textcolor{LimeGreen}{$\mathbf{y}_G$} (in our case -- termination
        probability) directly from the latent state
        \textcolor{Cyan}{$\mathbf{h}_G$} -- graph-level concepts are extracted
        through a full enumeration approach over the node concepts.
    }\label{fig:graph_conv_with_explanations}
\end{figure}

\paragraph{Encode-process-decode} Following the ``blueprint'' for neural
execution outlined in \citet{Velickovic2020Neural}, we model the algorithms by the
encode-process-decode architecture \citep{Hamrick2018Relational}. For each
algorithm $A$, an \emph{encoder network} $f_A$ encodes the algorithm-specific
node-level inputs $\mathbf{z}_i^{(t)}$ into the latent space. These node embeddings
are then processed using the \emph{processor network} $P$, usually a GNN. The
processor takes as input the encoded inputs
$\mathbf{Z}^{(t)}=\{\mathbf{z}_i^{(t)}\}_{i\in V}$ and graph edge index $E$ to
produce latent features $\mathbf{H}^{(t)}=\{\mathbf{h}_i^{(t)}\in
\mathbb{R}^{|\mathcal{L}|}\}_{i\in V}$, where $|\mathcal{L}|$ is the size of the
latent dimension. In contrast with previous work, we calculate algorithm
outputs by first passing the latent embeddings through a \emph{decoder network}
$g'_A$, which produces concepts for each node
$\mathbf{C}^{(t)}=\{\mathbf{c}_i^{(t)} \in (0,1)^{|\mathcal{C}|}\}$, where
$|\mathcal{C}|$ is number of concepts. The concepts are then passed through
a \emph{concept decoder} $g_A$ to produce node-level outputs
$\mathbf{Y}^{(t)}=\{\mathbf{y}_i^{(t)}\}$. 

Where applicable, we also
utilise a \emph{termination network} $T_A$ for deciding when to stop. However, in contrast with
prior work, we observed that training is more stable if we
calculate the termination probability based on potential \emph{next step}
embeddings (i.e.  a belief about what is the state after an iteration has been
executed). Additionally we found it insufficient to use the average node
embeddings as input to $T_A$ --  averaging would obfuscate the signal if
there is just a single node which should tell us whether to continue iterating
or not. Instead, we opted to use the output of an adapted PrediNet
\citep{Shanahan2020AnExplicitly} architecture with one attention head. PrediNet
is designed to represent the conjunction of elementary propositions, therefore
it can (theoretically) capture the logical bias of the termination rules. The
whole process is summarised in Figure \ref{fig:graph_conv_with_explanations} as
well as in the equations below:\\
{\centering
\begin{tabular}{p{6cm} p{6cm}}
{\begin{align}
    \label{eq:1}
    \mathbf{z}_i^{(t)} &= f_A\left(\mathbf{x}_i^{(t)}, \mathbf{h}_i^{(t-1)}\right)\\
    \mathbf{H}^{(t)} &= P\left(\mathbf{Z}^{(t)}, E\right)\\
    \mathbf{c}_i^{(t)} &= \sigma\left(g'_A\left(\mathbf{z}_i^{(t)}, \mathbf{h}_i^{(t)}\right)\right)\\
    \label{eq:4}\mathbf{y}_i^{(t)} &= g_A\left(\mathbf{c}_i^{(t)}\right)
\end{align}} & {\begin{align}
    \label{eq:5}
    \mathbf{z'}_i^{(t)} &= f_A\left(\mathbf{y}_i^{(t)}, \mathbf{h}_i^{(t)}\right)\\
    \mathbf{H'}^{(t)} &= P\left(\mathbf{Z'}^{(t)}, E\right)\\
    \overline{\mathbf{H}^{(t)}} &= \text{PrediNet}(\mathbf{H'}^{(t)})\\
    \label{eq:8}
    \tau^{(t)} &= \sigma\left(T_A\left(\overline{\mathbf{H}^{(t)}}\right)\right)
\end{align}} \\
\end{tabular}
}\\
where $\sigma$ is a logistic sigmoid function. When using $T_A$, equations
\ref{eq:1}-\ref{eq:8} are repeated if $\tau^{(t)} > 0.5$.

The combination of encoded inputs, together with the latent state of the given
node, contains sufficient information not only about the output at a given step,
but also: (i) a node's current state and (ii) observations about other nodes'
states in its neighbourhood. If our concepts are engineered to capture some
knowledge of either (i) or (ii), then we can extract meaningful algorithm
output explanations \emph{without providing any explicit information about
how the algorithm works (theorems, invariants, etc.)}

\paragraph{Explicitly relational GNN architecture} Some graph level
tasks (e.g. deciding termination) can be reduced to a logical formula over all
nodes -- for the algorithms and concepts we consider, termination can be
reduced to existence of a node with specific properties. (See graph-level rule
extraction). We engineer this logical bias into the termination network $\tau$
by adapting PrediNet \citep{Shanahan2020AnExplicitly} to the graph domain. The
PrediNet network architecture learns to represent conjunction/disjunction of
elementary propositions and is therefore suitable for the termination task. We
list the two minor modifications we made to adapt PrediNet to our tasks in
Appendix \ref{app:predinet}.



\paragraph{Extracting node-level algorithm rules} Deciding node-level formulas for
algorithm $A$ is achieved by examining the weights of the concept decoder
$g_A$. To achieve this, we used the open-source package 
\texttt{logic\_explained\_networks}\footnote{Apache 2.0 Licence.} \citep{barbiero2021lens}
implementing a wide collection of techniques to extract logic-based
explanations from concept-bottleneck neural networks
\citep{gori2017machine,ciravegna2020constraint}. The library takes as inputs
(i) node-level output decoder weights, (ii) predicted concepts from training
data, and (iii) training data ground truth labels, and generates logic formulas
in disjunctive normal form as outputs \citep{mendelson2009introduction}. By
construction, the concept decoder $\bar{g}_A \approx g_A$ learnt from concepts
$\mathcal{C}$ to outputs $\mathcal{O}$ is a Boolean map.  As any Boolean
function, it can be converted into a logic formula in disjunctive normal form
by means of its truth-table \citep{mendelson2009introduction}. The weights of
the concept decoder $g_A$ are used to select the most relevant concepts for
each output task. To get concise logic explanations \emph{when many concepts
are required as inputs} (in our experiments this is only the graph coloring
task), we add a regularization term in the loss function minimising the
$L1$-norm of the concept decoder weights $W$, leading to sparse
configurations of $W$. Later, at the training epoch $t_{\text{prune}}$,
first-layer weights are pruned concept-wise, i.e. removing \textit{all} the
weights departing from the least relevant concepts:
\begin{equation}
    \tilde{W}^1_j = W^1_j \mathbb{I}_{||W^1_j||_1 \geq \max_i ||W^1_i||_1/2}, \qquad\text{for } i=1,\dots,|\mathcal{C}|
\end{equation}
where $\mathbb{I}$ is the indicator function and $W^1$ are weights of the first
layer. Further details on logic extraction are provided in Appendix
\ref{app:formulas}.


\paragraph{Extracting algorithm termination rules} When deciding whether to
continue execution, we use the fact that a number of graph algorithms continue
iterating until a node with a specific combination of concepts exists. Since it
is unclear what combination of concepts we should supervise towards, we took a full enumeration approach when extracting rules
for termination. First, we generate a sample $j$ of the form $(\mathcal{U}_j, \tau'_j)$ from the training set from each step for a given graph.  $\tau'_j$ is
the ground truth for whether we should keep iterating, and
$\mathcal{U}_j=\{\mathbf{c'}_1, \dots, \mathbf{c'}_k\}$ is a set of all unique
concepts combinations,\footnote{$k$ may vary across samples} \emph{after the
algorithm update state has been performed} (hence $\mathbf{c'}$). Given a set
of concept indexes\footnote{$\{a..b\}$ denotes the set of integers from $a$ to
$b$} $\mathcal{I}\subseteq\mathcal{P}(\{1..|\mathcal{C}|\})$ and truth
assignment $\mathcal{T}: \{1..|\mathcal{C}|\}\to\{0, 1\}$ telling us which concepts
\emph{must} be true/false, we check if the following is satisfied:
\begin{equation}
    \forall j \left(\tau'_j=1 \iff \left(\exists \mathbf{c} \in
        \mathcal{U}_j.\ \forall \mathcal{I}_i \in \mathcal{I}.\
\mathbf{c}_{\mathcal{I}_i}=\mathcal{T}\left(\mathcal{I}_i\right)\right)\right)
\end{equation}
i.e. we should continue iterating if a special concept combination exists, and
we should stop iterating if it does not. We employ a brute-force approach for
finding $\mathcal{I}$ and $\mathcal{T}$, breaking ties by preferring smaller
$\mathcal{I}$.\footnote{In our case this broke all ties, but, if necessary, one
can add tie-break on truth assignment, by e.g. choosing assignments with
more true/false values} The complexity of such an approach is exponential, but
if the concept bottleneck is carefully engineered, the number of necessary
concepts, and/or the number of concepts in the special combination will be
small, making the computation feasible.

\emph{More importantly}, if the same enumeration approach was applied to the
raw node input data a $\mathcal{I}$/$\mathcal{T}$ combination \emph{may not
exist}. For example, the node-level inputs for the BFS task on each step do not
tell us which nodes have visited neigbours (crucial for deciding termination).
Additionally, if we have larger number of input features, the brute-force
approach \emph{may not be computationally feasible} -- the combinations scale
exponentially with the number of node features and concepts are one way to
reduce this number.

\section{Experimental setup}

The code for our experiments can be found at \url{https://github.com/HekpoMaH/algorithmic-concepts-reasoning}.

\paragraph{Algorithms considered} We apply our GNN to the following algorithms:
\emph{breadth-first search} (BFS), \emph{parallel coloring}
\citep{Jones1993APG}, a graph coloring heuristic, and \emph{Kruskal's
minimum spanning tree} (MST) algorithm  \citep{Kruskal1956Shortest}. BFS
is modelled as a binary classification problem where we predict whether a node is
visited or not, parallel coloring -- as a classification task over the classes
of possible node colors, plus one class for uncolored nodes. Kruskal's is
modelled as two tasks trained in parallel -- one, as a classification task to
choose the next edge to be considered for the MST and one to help us decide
which nodes belong to the same set. As the original Kruskal's algorithm
executes for $|E|$ steps, \emph{we do not learn termination for the MST task}
and stick to a fixed number of steps. We show how we model inputs/outputs for
all algorithms in Appendix \ref{app:modelling}.

\begin{table}
    \centering
    \caption{
        Algorithms and their corresponding concepts. We provide some sample
        ground truth explanations. Visual examples of how the algorithms work
        can be seen in Figure \ref{fig:visual_abstract}.
    }\label{tab:concepts}
    \begin{tabular}{l l l}
        \toprule
        \multirow{2}{*}{\textbf{Algorithm}} & \multirow{2}{*}{\textbf{Concepts}} & \textbf{Example ground-truth explanations}\\
                                            &  & (not provided to the model)\\
        \midrule
        \multirow{2}{*}{BFS} & $hasBeenVisited$ ($hBV$) & $hVN(i)\implies y_i^{(t)}=1$\\
                             & $hasVisitedNeighbours$ ($hVN$) & $\exists i. \lnot hBV(i)\land hVN(i) \implies \tau^{(t)}=1$\\
        \midrule
        \multirow{4}{*}{Coloring} &  & $iC(i)\land c1S(i) \land \lnot c2S(i) \implies y_i^{(t)}=2$ \\
                                  & \multirow{1}{*}{$isColored$ ($iC$), $hasPriority$ ($hP$)} &\\
                                  & \multirow{1}{*}{$colorXSeen$ ($cXS$), $X\in\{1,..,5\}$} &$(\lnot iC(i) \land hP(i)\land c1S(i) \land c2S(i)$\\
                                  & & $\land \ \lnot c3S(i)) \implies y_i^{(t)}=3$\\
        \midrule
        \multirow{5}{*}{Kruskal's} & & $(lEV(i)\land \lnot nISS(i) \land \lnot eIM(i))$ \\
                                  & $lighterEdgesVisited$ ($lEV$) & $ \implies y_i^{(t)}=1$ \\
                                  & \multirow{1}{*}{$nodesInSameSet$ ($nISS$)} &\\
                                  & \multirow{1}{*}{$edgeInMst$ ($eIM$)} & $(nISS(i) \land \lnot eIM(i))$\\
                                  & & $\implies y_i^{(t)}=0$\\
        \bottomrule
    \end{tabular}
\end{table}

Importantly, all algorithms we experimented with posses the following two
properties: (i) node/edge outputs are discrete and can be described in terms of
concepts; (ii) continuing the execution can be reduced to the existence of
a node with a specific combination of features. Examples of classical
algorithms that \emph{do not} fall into this category are the class of shortest
path algorithms: to explain such algorithms, we would need to use
arithmetic (e.g. minimum, sum) for the rules -- something that concepts cannot
directly capture. We leave explanation of such algorithms for future work.

To generate our concepts, we took into account what properties of the
nodes/neighbourhood the algorithm uses, but we did not provide any details to
the model \emph{how} to use them. Table \ref{tab:concepts} gives more details on
what concepts we chose and some example explanations. We give an outline how
one can use these concepts for explaining the algorithms, in Appendix
\ref{app:algosandconcepts}.




\paragraph{Data generation}
Following prior research on the topic of neural execution
\citep{Velickovic2020Neural}, for BFS we generate graphs from a number of
categories -- ladder, grid, Erd\H{o}s-R\'enyi \citep{Erdos1960Evolution},
Barab\'asi-Albert \citep{Albert2002Statistical}, 4-Community graphs, 4-Caveman
graphs and trees. For the coloring task, we limit the number of colors to 5 and
then generate graphs where \emph{all} nodes have fixed degree 5. This made the
task both challenging (i.e. there are occassions where 5 colors are necessary)
and feasible (we can generate graphs that are 5-colorable). Training data for
these tasks graph size is fixed at 20 and we test with graph sizes of 20, 50
and 100 nodes. For Kruskal's algorithm, we reused most graph categories for the
BFS task, except the last three where the graph is either a tree or is not
connected. Due to GPU memory constraints, training MST on graphs of size 20
required reducing the batch size by a factor of 4 and making the training very
time consuming. Therefore, for the MST task we reduced the size of the training
graphs to 8. Testing is still performed on graphs of size 20, 50 and 100.  For
all tasks we did a 10:1:1 train:validation:testing
split.\footnote{When working with multiple graph categories, the ratio is
preserved across each category} More details about the data generation are
present in Appendix \ref{app:data}.

\paragraph{Architectures tested} We decided to choose message-passing neural
networks \citep{Gilmer2017Neural} with the max aggregator for the main skeleton
of our processor (GNN) architecture as this type of GNN is known to align well
with algorithmic execution \citep{Velickovic2020Neural, Georgiev2020Neural,
Velickovic2020Pointer}. However, due to the nonlinear nature of some of the
tasks (parallel coloring) and the added concept bottleneck we found it
beneficial to add a hidden layer to some of the encoders and decoders, rather
than simply model them as an affine projection.

The Kruskal's algorithm consists of several steps -- masking out
visited edges, finding the minimal edge from the unmasked and checking if two
nodes are in the same set and unifying if they are not. The architecture for
this algorithm, follows the main ideas of Figures
\ref{fig:visual_abstract}\&\ref{fig:graph_conv_with_explanations}, to implement
them we combine the architecture of \citet{Yan2020Neural} for the first two
steps and \citet{Velickovic2020Pointer} for the third step. More details can be
found in Appendix \ref{app:implementingkruskal}.


\paragraph{Experimental details} We train our models using teacher forcing
\citep{Williams1989Learning} for a fixed number of epochs (500 for BFS, 3000
for the parallel coloring, 100 for Kruskal's). When testing BFS/parallel
coloring, we pick the model with the lowest sum of validation losses and when
testing Kruskal's -- the model with highest last-step accuracy. For training we
use Adam optimizer \citep{Kingma2015Adam} with initial learning rate of $0.001$
and batch size 32.  We optimise the sum of losses on the concept, output and
termination (except for Kruskal's, see above) predictions -- for more details
on how we define our losses see Appendix \ref{app:algosandlosses}. We evaluate
the ability to strongly generalise on graphs with sizes 50 and 100. Standard
deviations are obtained over 5 runs. For parallel coloring we add L1
regularisation and pruning on epoch 2000 to obtain higher quality explanations
since every combination of (concepts, output) pair may not be observed during
training. Libraries, code, and computing details are described in Appendix
\ref{app:software}. All hyperparameters were tuned manually.


\paragraph{Metrics} We use a variety of metrics, such as \emph{mean-step
accuracy} (average accuracy of per-step outputs), \emph{last-step accuracy}
(average accuracy of final algorithm outputs) and \emph{termination accuracy}
(average accuracy of predicting termination). Similarly, we define:
\emph{concepts mean-step accuracy} and \emph{concepts last-step accuracy} as
well \emph{formula mean-step accuracy}, \emph{formula last-step accuracy} and
\emph{formula termination accuracy}. The last three are derived by applying the
\emph{extracted formulas} to the \emph{predicted concepts} for predicting the
output/termination instead of using the respective
neural network. The motivation behind is that if we achieve high concept
accuracies and high formula accuracies then the formulas are likely to be
representing the underlying algorithm (or data) accurately.



\paragraph{Qualitative analysis} We provide several qualitative experiments:
(i) We fit a decision tree (DT) for the \CTO task (\CTT is not possible, due
to DTs working on fixed size node-level features). Concepts and targets are
obtained from the \emph{ground truth} concepts and target classes of \emph{all
training data nodes} at each step for each graph. (ii) We also plot the
concepts last/mean step accuracy vs epoch for each concept and provide further
analysis on which concept the networks find the most difficult. (iii) We
provide sample target class explanations for each algorithm.

\section{Results and discussion} \label{sec:results}

\begin{table}[t]
    \centering
    \caption{Parallel coloring accuracies over 5 runs}\label{tab:PCACC}
    \begin{tabular}{l l r r r}
        \toprule
        \textbf{Model} & \textbf{Metric} & \multicolumn{1}{c}{$|V|=20$} & \multicolumn{1}{c}{$|V|=50$} & \multicolumn{1}{c}{$|V|=100$}\\
        \midrule
         \multirow{3}{*}{\shortstack[l]{Standard}}
& mean-step acc. & $99.09{\scriptstyle\pm 0.86\%}$ & $98.74{\scriptstyle\pm 0.44\%}$ & $97.92{\scriptstyle\pm 1.50\%}$ \\
& last-step acc. & $99.25{\scriptstyle\pm 0.56\%}$ & $99.17{\scriptstyle\pm 0.20\%}$ & $99.13{\scriptstyle\pm 0.29\%}$ \\
& term. acc. & $98.79{\scriptstyle\pm 0.86\%}$ & $96.79{\scriptstyle\pm 1.53\%}$ & $95.08{\scriptstyle\pm 2.89\%}$ \\

                       \\
       \multirow{7}{*}{\shortstack[l]{Bottleneck\\(+L1 and prune)}}
& mean-step acc. & $99.71{\scriptstyle\pm 0.11\%}$ & $99.23{\scriptstyle\pm 0.21\%}$ & $98.92{\scriptstyle\pm 0.59\%}$ \\
& last-step acc. & $99.69{\scriptstyle\pm 0.13\%}$ & $99.17{\scriptstyle\pm 0.23\%}$ & $99.10{\scriptstyle\pm 0.22\%}$ \\
& term. acc. & $99.61{\scriptstyle\pm 0.18\%}$ & $99.02{\scriptstyle\pm 0.43\%}$ & $98.59{\scriptstyle\pm 0.77\%}$ \\
& formula mean-step acc. & $99.71{\scriptstyle\pm 0.12\%}$ & $99.24{\scriptstyle\pm 0.21\%}$ & $98.93{\scriptstyle\pm 0.59\%}$ \\
& formula last-step acc. & $99.69{\scriptstyle\pm 0.13\%}$ & $99.16{\scriptstyle\pm 0.22\%}$ & $99.08{\scriptstyle\pm 0.19\%}$ \\
& formula term. acc. & $99.51{\scriptstyle\pm 0.17\%}$ & $99.02{\scriptstyle\pm 0.43\%}$ & $98.48{\scriptstyle\pm 0.74\%}$ \\
& *concepts mean-step acc. & $99.85{\scriptstyle\pm 0.05\%}$ & $99.60{\scriptstyle\pm 0.10\%}$ & $99.45{\scriptstyle\pm 0.29\%}$ \\
& *concepts last-step acc. & $99.72{\scriptstyle\pm 0.07\%}$ & $99.35{\scriptstyle\pm 0.23\%}$ & $99.42{\scriptstyle\pm 0.09\%}$ \\



        \bottomrule
    \end{tabular}
\end{table}

\paragraph{Concept accuracies} As can be seen from Tables
\ref{tab:PCACC}\&\ref{tab:kruskrusACC} and Table \ref{tab:BFSACC}, Appendix
\ref{app:BFSACC} (metrics with an asterisk) we are able to learn concepts with
high accuracy (99\% and higher accuracy for BFS and
parallel coloring). Results show that GNNs are capable of producing high-level
concepts, capturing either node or neighbourhood information, for these
algorithmic tasks and the learned concept extractors strongly generalise --
concept accuracy does not drop even for $5\times$ larger graphs.

\paragraph{Parallel algorithms: BFS and coloring} For the BFS task both the
baseline and bottlenecked model perform optimally in line with the state of the
art. We therefore present results from the BFS task in Appendix
\ref{app:BFSACC}. Results from the parallel coloring task are shown in Table
\ref{tab:PCACC}. Apart from the high accuracy achieved, our results show that:
(i) the bottleneck doesn't have a major impact on the final model accuracy --
original metrics\footnote{namely mean-, last-step accuracy and termination
accuracy} remain the same or are better for both algorithms; (ii) we are able
to learn concepts accurately and (iii) the extracted rules are accurate --
applying them to the accurately predicted concepts in order to produce output
has no significant negative effect on the predictive accuracy of our model --
formula based accuracies do not deviate more than 5-6\% than the
original metrics.

\begin{table}[t]
    \centering
    \caption{Kruskal's algorithm accuracies over 5 runs}\label{tab:kruskrusACC}
    \begin{tabular}{l l r r r}
        \toprule
        \textbf{Model} & \textbf{Metric} & \multicolumn{1}{c}{$|V|=20$} & \multicolumn{1}{c}{$|V|=50$} & \multicolumn{1}{c}{$|V|=100$}\\
        \midrule
         \multirow{2}{*}{\shortstack[l]{Standard}}
    & mean-step acc. & $96.75{\scriptstyle\pm 0.15\%}$ & $95.41{\scriptstyle\pm 0.09\%}$ & $94.68{\scriptstyle\pm 0.10\%}$ \\
    & last-step acc. & $93.70{\scriptstyle\pm 0.33\%}$ & $90.10{\scriptstyle\pm 2.80\%}$ & $86.69{\scriptstyle\pm 4.28\%}$ \\
                       \\
       \multirow{6}{*}{{Bottleneck}}
& mean-step acc. & $96.93{\scriptstyle\pm 0.13\%}$ & $95.86{\scriptstyle\pm 0.37\%}$ & $95.27{\scriptstyle\pm 0.59\%}$ \\
& last-step acc. & $94.00{\scriptstyle\pm 0.24\%}$ & $92.20{\scriptstyle\pm 0.52\%}$ & $91.29{\scriptstyle\pm 0.86\%}$ \\
& formula mean-step acc. & $96.79{\scriptstyle\pm 0.37\%}$ & $95.77{\scriptstyle\pm 0.31\%}$ & $95.25{\scriptstyle\pm 0.54\%}$ \\
& formula last-step acc. & $93.70{\scriptstyle\pm 0.71\%}$ & $91.92{\scriptstyle\pm 0.47\%}$ & $91.15{\scriptstyle\pm 0.60\%}$ \\
& *concepts mean-step acc. & $97.91{\scriptstyle\pm 0.08\%}$ & $97.21{\scriptstyle\pm 0.22\%}$ & $96.80{\scriptstyle\pm 0.35\%}$ \\
& *concepts last-step acc. & $99.56{\scriptstyle\pm 0.29\%}$ & $99.49{\scriptstyle\pm 0.49\%}$ & $97.09{\scriptstyle\pm 0.29\%}$ \\



        \bottomrule
    \end{tabular}
\end{table}

\paragraph{Qualitative analysis: decision trees} We visualise the fitted
decision trees (DTs) for each algorithm in Appendix \ref{app:DT-node}. In all
cases the logic of the DT follows the logic of the original algorithm.
Additionally, the leaf nodes of all decision trees contain samples from
a single class showing that concepts were capable of capturing the complexity
of the algorithm.

\begin{table}
    \centering
    \caption{
        Sample explanations for each algorithm \emph{obtained from the learned
        model}. $cXS$ denotes $colorXSeen$, $nISS$ is $nodesInSameSet$, $lEV$
        is $lighterEdgesVisited$, $eIM$ is $edgeInMst$.
    }\label{tab:explanations}
\begin{tabular}{l l l}
    \toprule
    \textbf{Algorithm} & \textbf{Thing to explain} & \textbf{Explanation}\\
    \midrule

    \multirow{2}{*}{BFS} &$n$ is visited & {\scriptsize $hasVisitedNeighbours(n)$} \\
                       &continue execution & {\scriptsize $\exists n. \lnot hasBeenVisited(n) \land hasVisitedNeighbours(n)$} \\
   \midrule
    \multirow{7}{*}{\shortstack[l]{parallel\\coloring}}  & \multirow{2}{*}{$n$ has color 2} & {\scriptsize $(isColored(n) \land \lnot hasPriority(n) \land c1S(n) \land \lnot c2S(n)) \lor$} \\
                                                         & & {\scriptsize $(hasPriority(n) \land c1S(n) \land \lnot c2S(n) \land \lnot isColored(n))$}  \\
     \\
                                                         &\multirow{4}{*}{$n$ has color 5} & {\scriptsize $(isColored(n) \land \lnot hasPriority(n) \land c1S(n)$}\\
                                                         & & {\scriptsize $\land c2S(n) \land c3S(n) \land c4S(n)) \lor$}\\
                                                         & & {\scriptsize $(hasPriority(n) \land c1S(n) \land c2S(n)$} \\
                                                         & & {\scriptsize $\land c3S(n) \land c4S(n) \land \lnot isColored(n))$} \\
                                         & continue execution & {\scriptsize $\exists n. \lnot isColored(n)$}\\
    \midrule
    \multirow{2}{*}{Kruskal's} & \multirow{1}{*}{$e$ not in MST} & {\scriptsize $(nISS(e) \land \lnot eIM(e)) \lor(\lnot lEV(e) \land \lnot eIM(e))$}\\
              & \multirow{1}{*}{$e$ in MST} & {\scriptsize $(lEV(e) \land nISS(e) \land eIM(e)) \lor (lEV(e) \land \lnot nISS(e) \land \lnot eIM(e))$}\\
    \bottomrule
\end{tabular}
\end{table}

\begin{figure}[t]
    \begin{subfigure}{0.49\textwidth}
        \includegraphics[width=\textwidth]{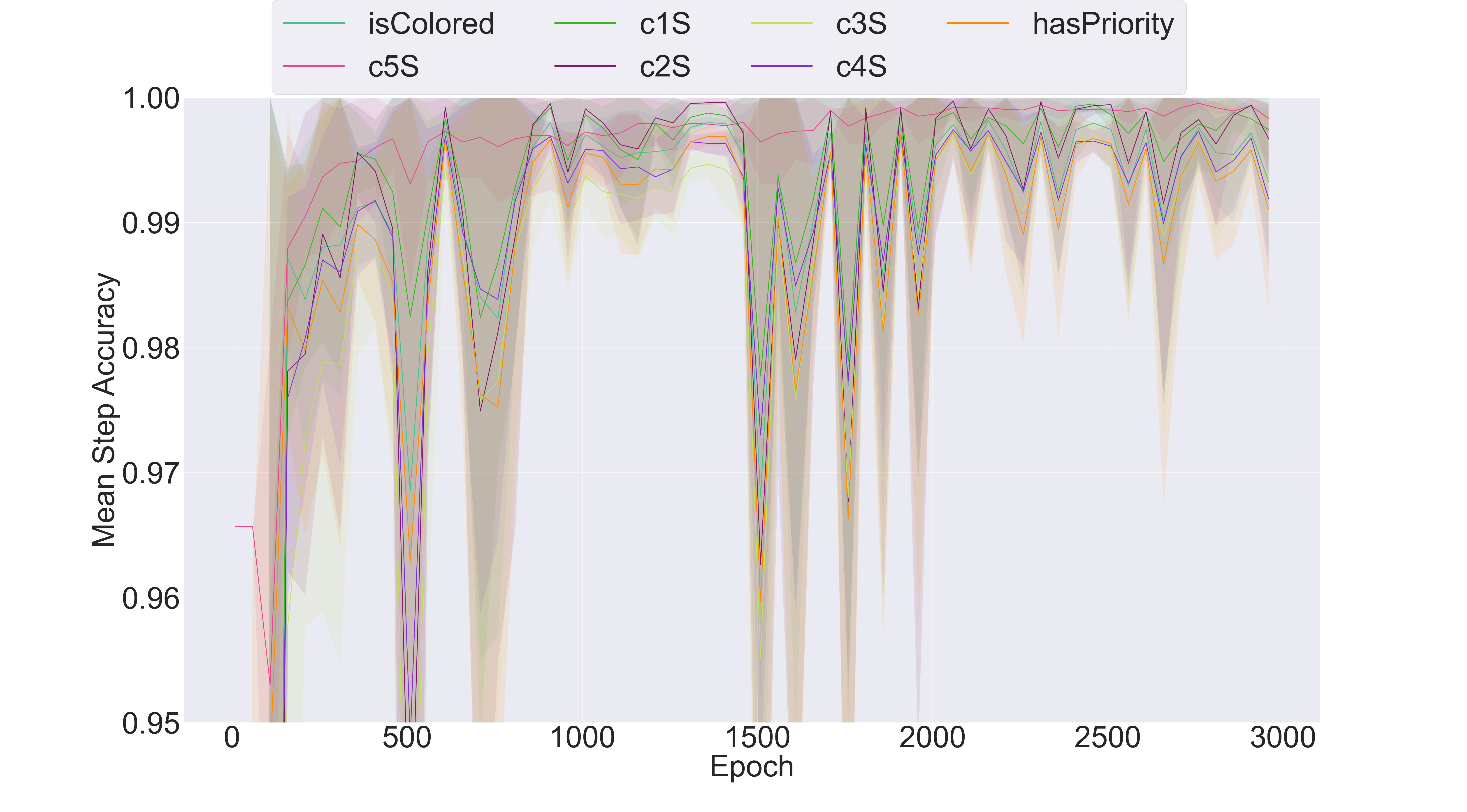}
        \caption{Concept mean-step accuracy}\label{fig:PCmsa}
    \end{subfigure}
    \begin{subfigure}{0.49\textwidth}
        \includegraphics[width=\textwidth]{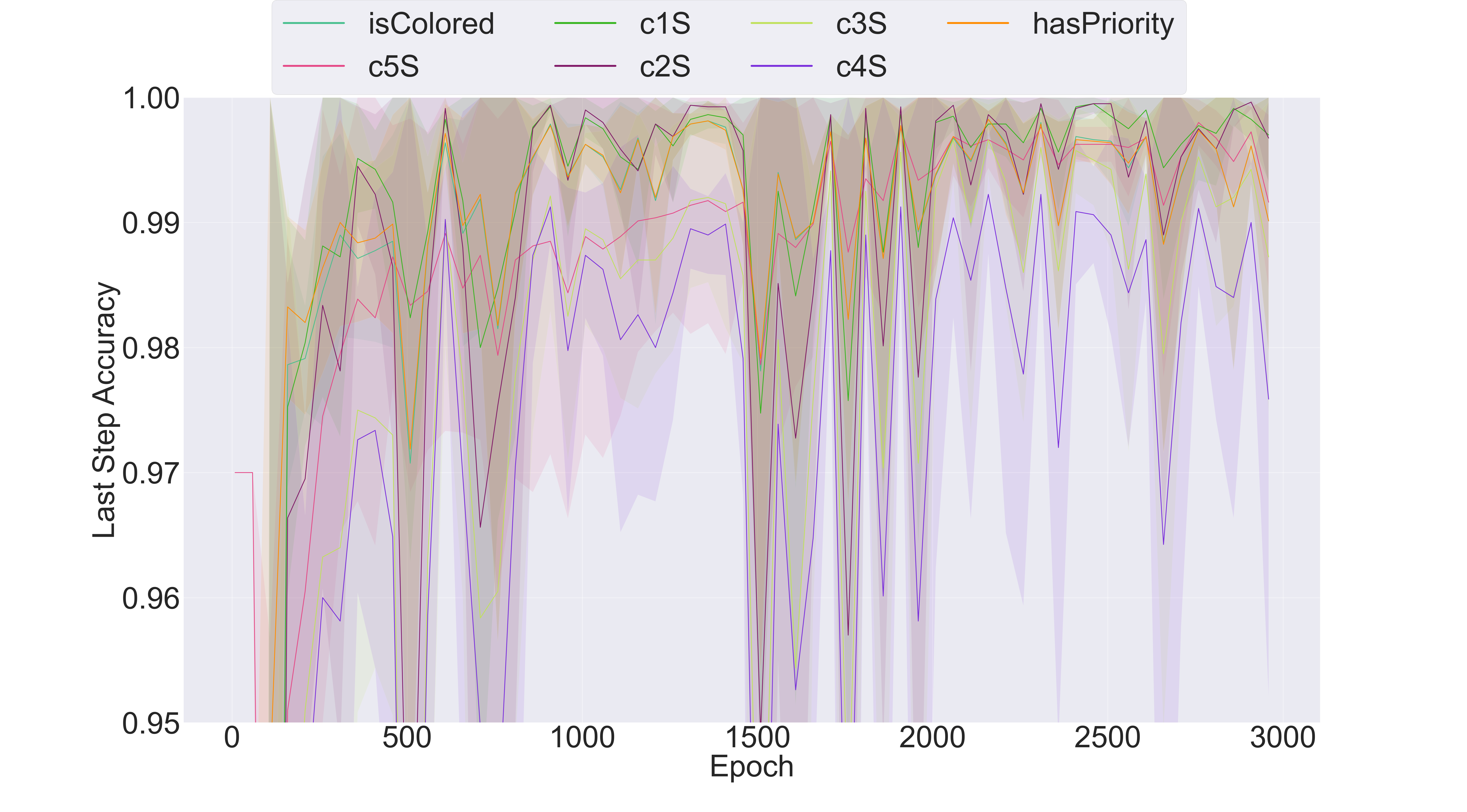}
        \caption{Concept last-step accuracy}
    \end{subfigure}
    \caption{
        Concept accuracies per epoch of the parallel coloring algorithm. (1
        point every 50 epochs). cXS is $colorXSeen$. Note the y axis scale. It
        can be observed that the $hasPriority$ concept is one of the worst
        performing concepts.  This leads to nodes being colored in a different
        order and therefore lower last-step concept accuracy for concepts
        related to colors. Standard deviation obtained from 5 runs.
    }\label{fig:concept_step_acc}
\end{figure}

\begin{figure}[t]
    \begin{subfigure}{0.49\textwidth}
        \includegraphics[width=\textwidth]{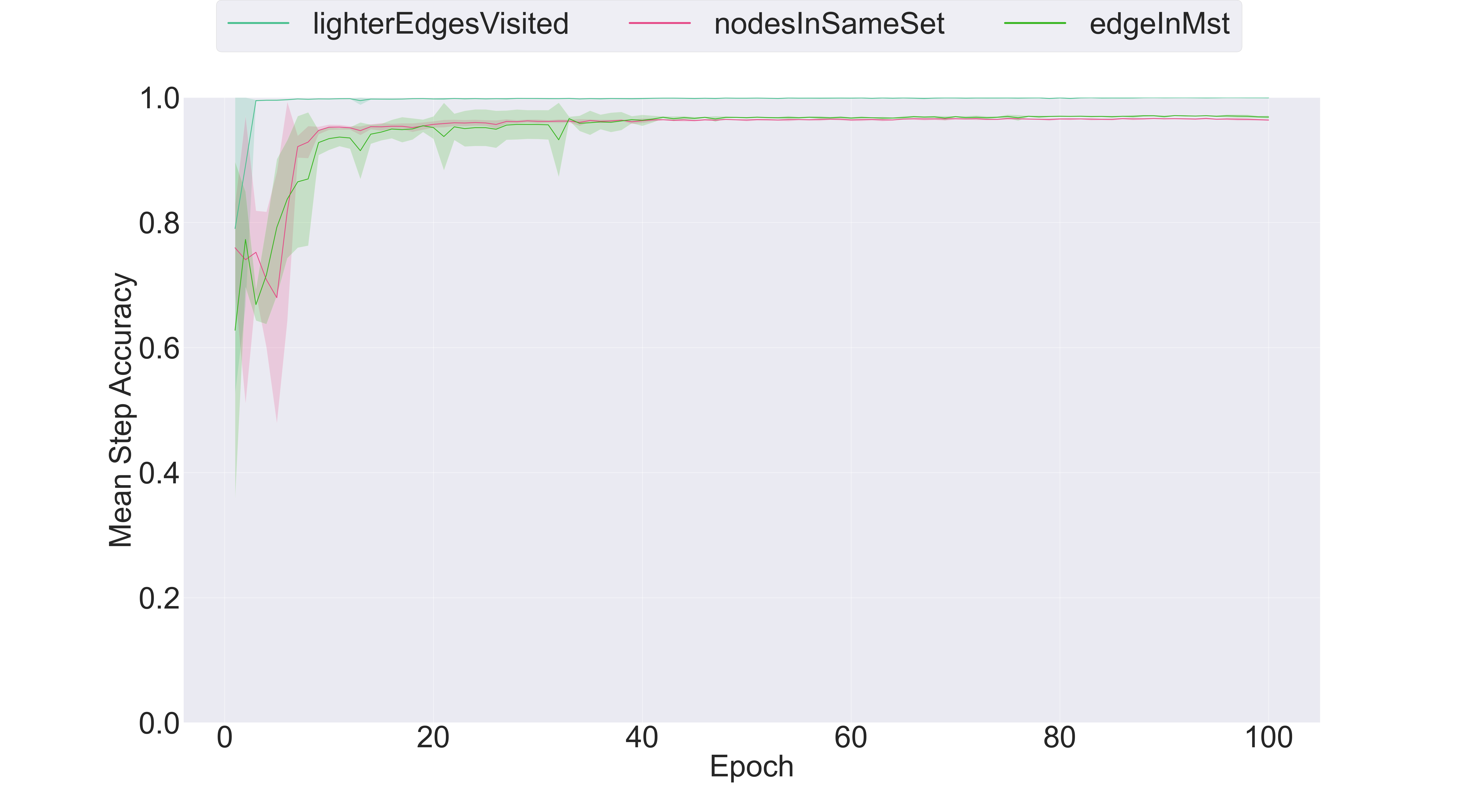}
        \caption{Concept mean-step accuracy}
    \end{subfigure}
    \begin{subfigure}{0.49\textwidth}
        \includegraphics[width=\textwidth]{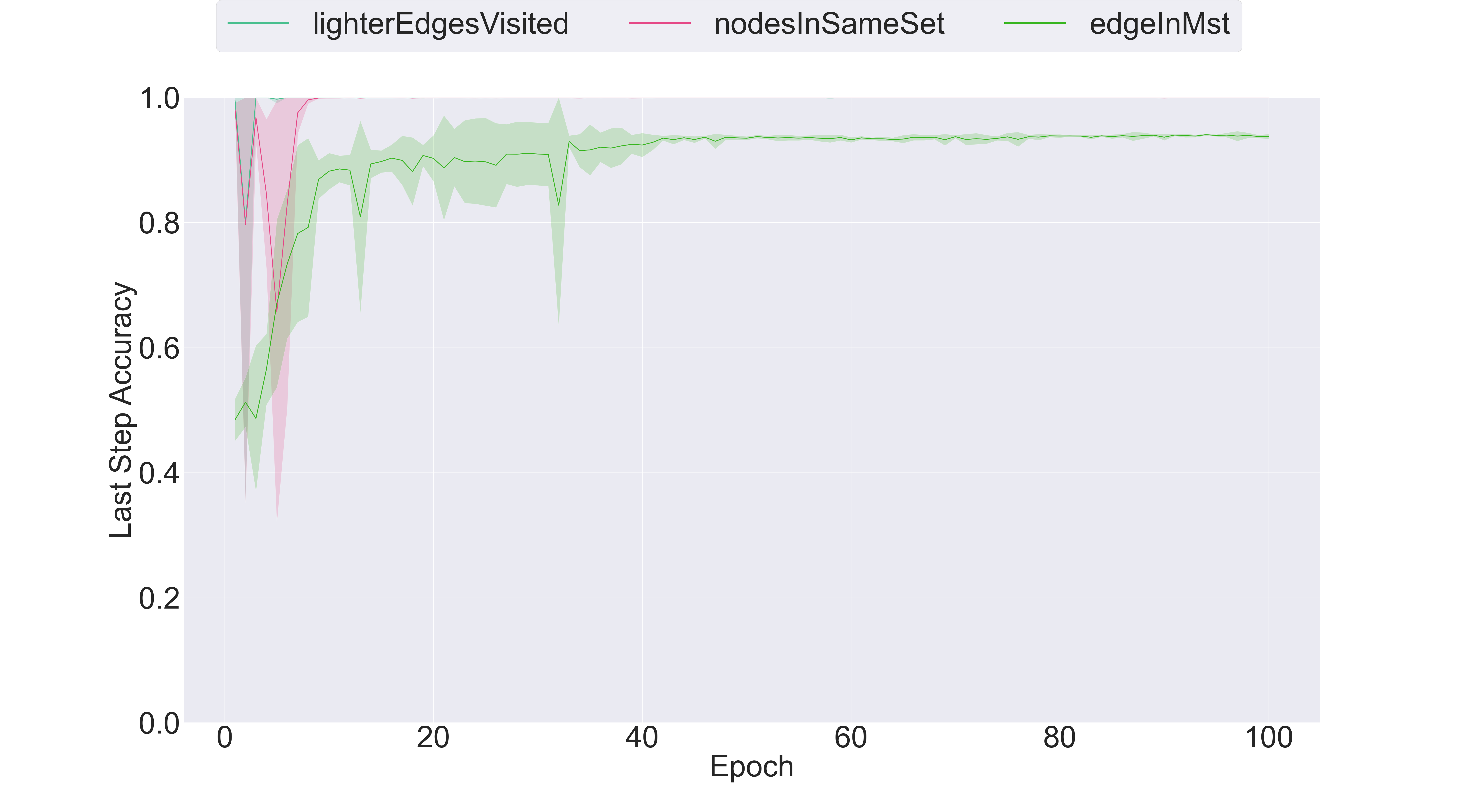}
        \caption{Concept last-step accuracy}
    \end{subfigure}
    \caption{
        Concept accuracies per epoch of the Kruskal's algorihm on graphs with
        20 nodes. (1 point per epoch). After an initial instability concepts
        are consistently accurate. Standard deviation obtained from 5 runs.
    }\label{fig:kruskal}
\end{figure}


\paragraph{Qualitative analysis: concept learning curves} We present per
concept learning curves for the parallel coloring in Figure
\ref{fig:concept_step_acc} and for Kruskal's in Figure \ref{fig:kruskal}: (i)
Parallel coloring exhibits many occasions where there are drops of concept
accuracy across almost all concepts. If we observe more carefully Figure
\ref{fig:PCmsa}, we will notice that they coincide with a drop of the accuracy
of $hasPriority$ concept. This drop also explains the lower last-step concept
accuracy -- changing the coloring order early on may produce quite different
final coloring. To confirm this observations, we trained an oracle model that
is always provided with the correct value for $hasPriority$. Such oracle model
achieved almost perfect concept accuracy -- we provide a plot of the concept
learning curves in Appendix \ref{app:oraclecoloring}; (ii) The concept
instability was present only in the beginning for Kruskal's, but it converged
to a stable solution. The reason $edgeInMst$ concept remained with the lowest
last-step accuracy is that the overall last-step accuracy of the model is lower.

\paragraph{Qualitative analysis: explanations} We list examples of obtained
explanations in Table \ref{tab:explanations} and present \emph{all}
explanations obtained from the algorithms in in Appendix
\ref{app:algo-explanations}. The extracted rules show that concepts are one way
to extract accurate representation of the rules of the algorithm. E.g. we can
(re)infer from the listed rules for parallel coloring that for getting a given
color that color should not be seen in the neighbourhood and colors coming
before that one have already been seen.

We additionally observed, that as the number of concepts increases, if we need
shorter and more general rules we need more and more data. One way to alleviate
such problem is L1 regularisation and pruning -- we additionally perform an
ablation study in Appendix \ref{app:algo-explanations} showing that without
regularisation rules are still usable (giving good formula accuracy) \emph{but
are less general}.


\section{Conclusions}

We presented concept-based reasoning on graph algorithms through Concept
Bottleneck Graph Neural Networks. We demonstrated through the surveyed
algorithms, that we can accurately learn node-level concepts without impacting
performance.
Moreover, by examining training data and model weights, we are capable of
explaining each node-level output classes with formulas based on the defined
concepts. Concepts also allow us perform a \emph{unsupervised} rule extraction
of certain graph-level tasks, such as deciding when to terminate. Extracted
rules are interpretable and applying them does not heavily impact accuracy.




\bibliographystyle{apalike}
\bibliography{refs}

\begin{thebibliography}{}

\bibitem[Adadi and Berrada, 2018]{adadi2018peeking}
Adadi, A. and Berrada, M. (2018).
\newblock Peeking inside the black-box: a survey on explainable artificial
  intelligence (xai).
\newblock {\em IEEE access}, 6:52138--52160.

\bibitem[Albert and Barab{\'a}si, 2002]{Albert2002Statistical}
Albert, R. and Barab{\'a}si, A.-L. (2002).
\newblock Statistical mechanics of complex networks.
\newblock {\em Reviews of modern physics}, 74(1):47.

\bibitem[Baldassarre and Azizpour, 2019]{baldassarre2019explainability}
Baldassarre, F. and Azizpour, H. (2019).
\newblock Explainability techniques for graph convolutional networks.
\newblock {\em arXiv preprint arXiv:1905.13686}.

\bibitem[Barbiero et~al., 2021]{barbiero2021lens}
Barbiero, P., Ciravegna, G., Georgiev, D., and Giannini, F. (2021).
\newblock Lens: a python library for logic explained networks.
\newblock arXiv preprint.

\bibitem[Battaglia et~al., 2016]{Battaglia2016Interaction}
Battaglia, P.~W., Pascanu, R., Lai, M., Rezende, D.~J., and Kavukcuoglu, K.
  (2016).
\newblock Interaction networks for learning about objects, relations and
  physics.
\newblock In Lee, D.~D., Sugiyama, M., von Luxburg, U., Guyon, I., and Garnett,
  R., editors, {\em Advances in Neural Information Processing Systems 29:
  Annual Conference on Neural Information Processing Systems 2016, December
  5-10, 2016, Barcelona, Spain}, pages 4502--4510.

\bibitem[Cappart et~al., 2021]{Cappart2021Combinatorial}
Cappart, Q., Ch{\'e}telat, D., Khalil, E., Lodi, A., Morris, C., and
  Veli{\v{c}}kovi{\'c}, P. (2021).
\newblock Combinatorial optimization and reasoning with graph neural networks.
\newblock {\em arXiv preprint arXiv:2102.09544}.

\bibitem[Chen et~al., 2020]{chen2020concept}
Chen, Z., Bei, Y., and Rudin, C. (2020).
\newblock Concept whitening for interpretable image recognition.
\newblock {\em Nature Machine Intelligence}, 2(12):772--782.

\bibitem[Ciravegna et~al., 2020]{ciravegna2020constraint}
Ciravegna, G., Giannini, F., Melacci, S., Maggini, M., and Gori, M. (2020).
\newblock A constraint-based approach to learning and explanation.
\newblock In {\em AAAI}, pages 3658--3665.

\bibitem[Deac et~al., 2020]{Deac2020XLVIN}
Deac, A., Velickovic, P., Milinkovic, O., Bacon, P., Tang, J., and Nikolic, M.
  (2020).
\newblock {XLVIN:} executed latent value iteration nets.
\newblock {\em CoRR}, abs/2010.13146.

\bibitem[Doshi-Velez and Kim, 2017]{doshi2017towards}
Doshi-Velez, F. and Kim, B. (2017).
\newblock Towards a rigorous science of interpretable machine learning.
\newblock {\em arXiv preprint arXiv:1702.08608}.

\bibitem[Erd{\H{o}}s and R{\'e}nyi, 1960]{Erdos1960Evolution}
Erd{\H{o}}s, P. and R{\'e}nyi, A. (1960).
\newblock On the evolution of random graphs.
\newblock {\em Publ. Math. Inst. Hung. Acad. Sci}, 5(1):17--60.

\bibitem[Galler and Fisher, 1964]{galler1964improved}
Galler, B.~A. and Fisher, M.~J. (1964).
\newblock An improved equivalence algorithm.
\newblock {\em Communications of the ACM}, 7(5):301--303.

\bibitem[Georgiev and Li\`{o}, 2020]{Georgiev2020Neural}
Georgiev, D. and Li\`{o}, P. (2020).
\newblock Neural bipartite matching.
\newblock In {\em Graph Representation Learning and Beyond (GRL+) workshop}.

\bibitem[Ghorbani et~al., 2019]{ghorbani2019towards}
Ghorbani, A., Wexler, J., Zou, J., and Kim, B. (2019).
\newblock Towards automatic concept-based explanations.
\newblock {\em arXiv preprint arXiv:1902.03129}.

\bibitem[Gilmer et~al., 2017]{Gilmer2017Neural}
Gilmer, J., Schoenholz, S.~S., Riley, P.~F., Vinyals, O., and Dahl, G.~E.
  (2017).
\newblock Neural message passing for quantum chemistry.
\newblock In {\em Proceedings of the 34th International Conference on Machine
  Learning, {ICML} 2017, Sydney, NSW, Australia, 6-11 August 2017}, pages
  1263--1272.

\bibitem[Gori, 2017]{gori2017machine}
Gori, M. (2017).
\newblock {\em Machine Learning: A constraint-based approach}.
\newblock Morgan Kaufmann.

\bibitem[Hamrick et~al., 2018]{Hamrick2018Relational}
Hamrick, J.~B., Allen, K.~R., Bapst, V., Zhu, T., McKee, K.~R., Tenenbaum, J.,
  and Battaglia, P.~W. (2018).
\newblock Relational inductive bias for physical construction in humans and
  machines.
\newblock In {\em Proceedings of the 40th Annual Meeting of the Cognitive
  Science Society, CogSci 2018, Madison, WI, USA, July 25-28, 2018}.

\bibitem[Jones and Plassmann, 1993]{Jones1993APG}
Jones, M. and Plassmann, P. (1993).
\newblock A parallel graph coloring heuristic.
\newblock {\em SIAM J. Sci. Comput.}, 14:654--669.

\bibitem[Joshi et~al., 2020]{Joshi2020Learning}
Joshi, C.~K., Cappart, Q., Rousseau, L., Laurent, T., and Bresson, X. (2020).
\newblock Learning {TSP} requires rethinking generalization.
\newblock {\em CoRR}, abs/2006.07054.

\bibitem[Kazhdan et~al., 2020a]{kazhdan2020meme}
Kazhdan, D., Dimanov, B., Jamnik, M., and Li{\`o}, P. (2020a).
\newblock Meme: Generating rnn model explanations via model extraction.
\newblock {\em arXiv preprint arXiv:2012.06954}.

\bibitem[Kazhdan et~al., 2020b]{kazhdan2020now}
Kazhdan, D., Dimanov, B., Jamnik, M., Li{\`o}, P., and Weller, A. (2020b).
\newblock Now you see me (cme): Concept-based model extraction.
\newblock {\em arXiv preprint arXiv:2010.13233}.

\bibitem[Kazhdan et~al., 2021]{kazhdan2021disentanglement}
Kazhdan, D., Dimanov, B., Terre, H.~A., Jamnik, M., Li{\`o}, P., and Weller, A.
  (2021).
\newblock Is disentanglement all you need? comparing concept-based \&
  disentanglement approaches.
\newblock {\em arXiv preprint arXiv:2104.06917}.

\bibitem[Kingma and Ba, 2015]{Kingma2015Adam}
Kingma, D.~P. and Ba, J. (2015).
\newblock Adam: {A} method for stochastic optimization.
\newblock In {\em 3rd International Conference on Learning Representations,
  {ICLR} 2015, San Diego, CA, USA, May 7-9, 2015, Conference Track
  Proceedings}.

\bibitem[Koh et~al., 2020]{Koh2020Concept}
Koh, P.~W., Nguyen, T., Tang, Y.~S., Mussmann, S., Pierson, E., Kim, B., and
  Liang, P. (2020).
\newblock Concept bottleneck models.
\newblock In {\em Proceedings of the 37th International Conference on Machine
  Learning, {ICML} 2020, 13-18 July 2020, Virtual Event}, volume 119 of {\em
  Proceedings of Machine Learning Research}, pages 5338--5348. {PMLR}.

\bibitem[Kruskal, 1956]{Kruskal1956Shortest}
Kruskal, J.~B. (1956).
\newblock On the shortest spanning subtree of a graph and the traveling
  salesman problem.
\newblock {\em Proceedings of the American Mathematical society}, 7(1):48--50.

\bibitem[Luo et~al., 2020]{luo2020parameterized}
Luo, D., Cheng, W., Xu, D., Yu, W., Zong, B., Chen, H., and Zhang, X. (2020).
\newblock Parameterized explainer for graph neural network.
\newblock {\em arXiv preprint arXiv:2011.04573}.

\bibitem[McCluskey, 1956]{mccluskey1956minimization}
McCluskey, E.~J. (1956).
\newblock Minimization of boolean functions.
\newblock {\em The Bell System Technical Journal}, 35(6):1417--1444.

\bibitem[McColl, 1878]{mccoll1878calculus}
McColl, H. (1878).
\newblock The calculus of equivalent statements (third paper).
\newblock {\em Proceedings of the London Mathematical Society}, 1(1):16--28.

\bibitem[Mendelson, 2009]{mendelson2009introduction}
Mendelson, E. (2009).
\newblock {\em Introduction to mathematical logic}.
\newblock CRC press.

\bibitem[Molnar, 2020]{molnar2020interpretable}
Molnar, C. (2020).
\newblock {\em Interpretable machine learning}.
\newblock Lulu. com.

\bibitem[Moore, 1959]{moore1959shortest}
Moore, E.~F. (1959).
\newblock The shortest path through a maze.
\newblock In {\em Proc. Int. Symp. Switching Theory, 1959}, pages 285--292.

\bibitem[Pal et~al., 2020]{Pal2020PinnerSage}
Pal, A., Eksombatchai, C., Zhou, Y., Zhao, B., Rosenberg, C., and Leskovec, J.
  (2020).
\newblock Pinnersage: Multi-modal user embedding framework for recommendations
  at pinterest.
\newblock In Gupta, R., Liu, Y., Tang, J., and Prakash, B.~A., editors, {\em
  {KDD} '20: The 26th {ACM} {SIGKDD} Conference on Knowledge Discovery and Data
  Mining, Virtual Event, CA, USA, August 23-27, 2020}, pages 2311--2320. {ACM}.

\bibitem[Paszke et~al., 2019]{paszke2019pytorch}
Paszke, A., Gross, S., Massa, F., Lerer, A., Bradbury, J., Chanan, G., Killeen,
  T., Lin, Z., Gimelshein, N., Antiga, L., et~al. (2019).
\newblock Pytorch: An imperative style, high-performance deep learning library.
\newblock {\em arXiv preprint arXiv:1912.01703}.

\bibitem[Pedregosa et~al., 2011]{pedregosa2011scikit}
Pedregosa, F., Varoquaux, G., Gramfort, A., Michel, V., Thirion, B., Grisel,
  O., Blondel, M., Prettenhofer, P., Weiss, R., Dubourg, V., et~al. (2011).
\newblock Scikit-learn: Machine learning in python.
\newblock {\em the Journal of machine Learning research}, 12:2825--2830.

\bibitem[Pope et~al., 2019]{pope2019explainability}
Pope, P.~E., Kolouri, S., Rostami, M., Martin, C.~E., and Hoffmann, H. (2019).
\newblock Explainability methods for graph convolutional neural networks.
\newblock In {\em Proceedings of the IEEE/CVF Conference on Computer Vision and
  Pattern Recognition}, pages 10772--10781.

\bibitem[Pr{\"u}fer, 1918]{Prufer1918Neuer}
Pr{\"u}fer, H. (1918).
\newblock Neuer beweis eines satzes {\"u}ber permutationen.
\newblock {\em Arch. Math. Phys}, 27(1918):742--744.

\bibitem[Quine, 1952]{quine1952problem}
Quine, W.~V. (1952).
\newblock The problem of simplifying truth functions.
\newblock {\em The American mathematical monthly}, 59(8):521--531.

\bibitem[Schnake et~al., 2020]{schnake2020higher}
Schnake, T., Eberle, O., Lederer, J., Nakajima, S., Sch{\"u}tt, K., M{\"u}ller,
  K., and Montavon, G. (2020).
\newblock Higher-order explanations of graph neural networks via relevant
  walks.
\newblock {\em arXiv: 2006.03589}.

\bibitem[Shanahan et~al., 2020]{Shanahan2020AnExplicitly}
Shanahan, M., Nikiforou, K., Creswell, A., Kaplanis, C., Barrett, D. G.~T., and
  Garnelo, M. (2020).
\newblock An explicitly relational neural network architecture.
\newblock In {\em Proceedings of the 37th International Conference on Machine
  Learning, {ICML} 2020, 13-18 July 2020, Virtual Event}, volume 119 of {\em
  Proceedings of Machine Learning Research}, pages 8593--8603. {PMLR}.

\bibitem[Stokes et~al., 2020]{Stokes2020ADeep}
Stokes, J.~M., Yang, K., Swanson, K., Jin, W., Cubillos-Ruiz, A., Donghia,
  N.~M., MacNair, C.~R., French, S., Carfrae, L.~A., Bloom-Ackermann, Z., Tran,
  V.~M., Chiappino-Pepe, A., Badran, A.~H., Andrews, I.~W., Chory, E.~J.,
  Church, G.~M., Brown, E.~D., Jaakkola, T.~S., Barzilay, R., and Collins,
  J.~J. (2020).
\newblock {A Deep Learning Approach to Antibiotic Discovery}.
\newblock {\em Cell}, 180(4):688--702.e13.

\bibitem[Strathmann et~al., 2021]{Strathmann2021Persistent}
Strathmann, H., Barekatain, M., Blundell, C., and Veli{\v{c}}kovi{\'c}, P.
  (2021).
\newblock Persistent message passing.
\newblock In {\em ICLR 2021 Workshop on Geometrical and Topological
  Representation Learning}.

\bibitem[Tang et~al., 2020]{Tang2020Towards}
Tang, H., Huang, Z., Gu, J., Lu, B.-L., and Su, H. (2020).
\newblock Towards scale-invariant graph-related problem solving by iterative
  homogeneous gnns.
\newblock {\em Advances in Neural Information Processing Systems}, 33.

\bibitem[Veli\v{c}kovi\'{c} et~al., 2020a]{Velickovic2020Pointer}
Veli\v{c}kovi\'{c}, P., Buesing, L., Overlan, M.~C., Pascanu, R., Vinyals, O.,
  and Blundell, C. (2020a).
\newblock Pointer graph networks.
\newblock In Larochelle, H., Ranzato, M., Hadsell, R., Balcan, M., and Lin, H.,
  editors, {\em Advances in Neural Information Processing Systems 33: Annual
  Conference on Neural Information Processing Systems 2020, NeurIPS 2020,
  December 6-12, 2020, virtual}.

\bibitem[Veli\v{c}kovi\'{c} et~al., 2020b]{Velickovic2020Neural}
Veli\v{c}kovi\'{c}, P., Ying, R., Padovano, M., Hadsell, R., and Blundell, C.
  (2020b).
\newblock Neural execution of graph algorithms.
\newblock In {\em 8th International Conference on Learning Representations,
  {ICLR} 2020, Addis Ababa, Ethiopia, April 26-30, 2020}. OpenReview.net.

\bibitem[Vu and Thai, 2020]{vu2020pgm}
Vu, M.~N. and Thai, M.~T. (2020).
\newblock Pgm-explainer: Probabilistic graphical model explanations for graph
  neural networks.
\newblock {\em arXiv preprint arXiv:2010.05788}.

\bibitem[Watts, 1999]{Watts1999Networks}
Watts, D.~J. (1999).
\newblock Networks, dynamics, and the small-world phenomenon.
\newblock {\em American Journal of sociology}, 105(2):493--527.

\bibitem[Wijaya et~al., 2021]{wijaya2021failing}
Wijaya, M.~A., Kazhdan, D., Dimanov, B., and Jamnik, M. (2021).
\newblock Failing conceptually: Concept-based explanations of dataset shift.
\newblock {\em arXiv preprint arXiv:2104.08952}.

\bibitem[Williams and Zipser, 1989]{Williams1989Learning}
Williams, R.~J. and Zipser, D. (1989).
\newblock A learning algorithm for continually running fully recurrent neural
  networks.
\newblock {\em Neural computation}, 1(2):270--280.

\bibitem[Xu et~al., 2020]{Xu2020What}
Xu, K., Li, J., Zhang, M., Du, S.~S., Kawarabayashi, K., and Jegelka, S.
  (2020).
\newblock What can neural networks reason about?
\newblock In {\em 8th International Conference on Learning Representations,
  {ICLR} 2020, Addis Ababa, Ethiopia, April 26-30, 2020}. OpenReview.net.

\bibitem[Yan et~al., 2020]{Yan2020Neural}
Yan, Y., Swersky, K., Koutra, D., Ranganathan, P., and Hashemi, M. (2020).
\newblock Neural execution engines: Learning to execute subroutines.
\newblock In Larochelle, H., Ranzato, M., Hadsell, R., Balcan, M., and Lin, H.,
  editors, {\em Advances in Neural Information Processing Systems 33: Annual
  Conference on Neural Information Processing Systems 2020, NeurIPS 2020,
  December 6-12, 2020, virtual}.

\bibitem[Yeh et~al., 2019]{yeh2019completeness}
Yeh, C.-K., Kim, B., Arik, S.~O., Li, C.-L., Pfister, T., and Ravikumar, P.
  (2019).
\newblock On completeness-aware concept-based explanations in deep neural
  networks.
\newblock {\em arXiv preprint arXiv:1910.07969}.

\bibitem[Ying et~al., 2019]{ying2019gnnexplainer}
Ying, R., Bourgeois, D., You, J., Zitnik, M., and Leskovec, J. (2019).
\newblock Gnnexplainer: Generating explanations for graph neural networks.
\newblock {\em Advances in neural information processing systems}, 32:9240.

\end{thebibliography}

\clearpage

\appendix

\section{Adapting PrediNet to graphs}\label{app:predinet}

As already discussed, PrediNet \citep{Shanahan2020AnExplicitly} is an
architecture that operates on the image and learns to capture logical bias.
To adapt it to the graph domain and more specifically the domain of our tasks,
we made the following two modifications: (i) Instead of flattening a convoluted
image we use global graph pooling (max pooling) for the calculation of the
query and (ii) we do not concatenate any positional information to the input
feature vectors. Everything else is kept the same as the original PrediNet:
\begin{align}
    \mathbf{Q}^h_1 &= \text{pool}(\mathbf{H'}^{(t)})W^h_{Q1}\\
    \mathbf{Q}^h_2 &= \text{pool}(\mathbf{H'}^{(t)})W^h_{Q2}\\
    \mathbf{K} &= \mathbf{H'}^{(t)}W_k\\
    \mathbf{E}^h_1 &= \text{softmax}(\mathbf{Q}^h_1 K^\top)\mathbf{H'}^{(t)}\\
    \mathbf{E}^h_2 &= \text{softmax}(\mathbf{Q}^h_2 K^\top)\mathbf{H'}^{(t)}\\
    \mathbf{D}^h &= \mathbf{E}^h_1 W_S - \mathbf{E}^h_2 W_S\\
    \mathbf{out} &= \text{Linear}\left(\mathbf{D}^h\right)
\end{align}

\section{Extraction of formulas}
\label{app:formulas}

By construction, the function $\bar{f} \approx f$ learnt from concepts $\mathcal{C}$ to outputs $\mathcal{O}$ is a Boolean map.
As any Boolean function, the map $\bar{f}: \mathcal{C} \rightarrow \mathcal{O}$ can be converted into a First-Order Logic (FOL) formula $\varphi$ in Disjunctive Normal Form (DNF) \citep{mendelson2009introduction} by means of its truth-table. The truth table provides a formal mechanism to extract logic rules of increasing complexity for individual observations, for cluster of samples, or for a whole class. 

\paragraph{FOL extraction}
A logic formula can be extracted for each sample $c \in \mathcal{C}$ corresponding to a single row of the truth table:
\begin{equation}
    \varphi_{c} = \tilde{c}_1\wedge\ldots\wedge\tilde{c}_k \qquad\mbox{where } \tilde{c}_j:= 
    \begin{cases}
        \bar{c}_j, & \mbox{if }c_j \geq t\\
        \neg \bar{c}_j, &\mbox{if } c_j < t\\
    \end{cases}\ ,    
    \mbox{ for $j=1,\ldots,k$,}
    \label{eq:locexp}
\end{equation}

Example-level formulas can be formally aggregated providing logic rules $\varphi_{S}$ for a cluster of samples $S \subseteq S^* = \{ \bar{c} \ | \ \bar{f}(\bar{c}) = t \}$ of the same class $t$
\begin{equation}
    \varphi_{S} = \displaystyle\bigvee_{c \in S}\varphi_{c} = \displaystyle\bigvee_{c \in S}\tilde{c}_1\wedge\ldots\wedge\tilde{c}_k
\label{eq:agg_exp}
\end{equation}

Any formula $\varphi_{S}$ can be thought of as a $\{0,1\}$-valued mapping defined on $\mathcal{C}$, that is equal to 1 exactly on a cluster $S$. We may also get an explicit explanation relating $\varphi_{S}$ and the Boolean function $\bar{f}$ by the FOL formula:
\[
\forall \bar{c}\in S:\ \varphi_{S}(\bar{c})\rightarrow \bar{f}(\bar{c})
\]

A formula for a whole class $t$ (e.g. $t=1$) can be formally obtained using Eq. \ref{eq:agg_exp} by considering $S^*$, i.e. aggregating all the example-level rules for the same class $t$. Also in this case it is possible to get a FOL formula relating $\varphi$ and the Boolean map $\bar{f}$ by means of:
\[
\forall \bar{c} \in S^*:\ \bar{f}(\bar{c})\leftrightarrow\varphi(\bar{c})
\]

\paragraph{FOL simplification} The aggregation of many example-level explanations may increase the length and complexity of the FOL formula being extracted for a cluster of samples or for a whole class. However, existing techniques as the Quine–McCluskey algorithm can be used to get compact and simplified equivalent FOL expressions \citep{mccoll1878calculus,quine1952problem,mccluskey1956minimization}. For instance, the explanation "(\textit{person} $\wedge$ \textit{nose}) $\vee$ ($\neg$\textit{person} $\wedge$ \textit{nose})" can be formally simplified in "\textit{nose}".

\section{Modelling algorithmic reasoning}\label{app:modelling}

We begin the modelling by designing the initial states for the algorithm:
\begin{align}
    \text{BFS} : x_i^{(1)} =
    \begin{cases}
        1 & i=s \\
        1 & i\neq s \\
        0 & \text{otherwise}
    \end{cases}
    \quad
    \text{coloring} : x_i^{(1)} = (0, binary(p_i))
    \quad
    \text{Kruskal's} : x_i^{(1)} = (0, w_i)
\end{align}
where $s$ is a randomly chosen starting node, $p_i$ denotes a randomly
chosen\footnote{We ensure no two nodes connected by an edge have the same
priority} \emph{constant} integer priority for node $i$ and $w_i$ denotes the
weight of edge $i$. For parallel coloring we leave class 0 for denoting
uncolored nodes in the coloring task. The extra 0 in the input for Kruskal's will
be a bit value whether an edge is selected to be in the minimum spanning tree.
Similar to \citet{Yan2020Neural} we represent numbers in binary, and learn an
embedding for each bit position. We also one hot encode any class information,
such as visited/unvisited or color of a node (or no color present).

Next step inputs are calculated according to the specific algorithm: for BFS,
a node becomes reachable if it had visited neighbours\footnote{As we utilise
self-loops for retention of self-information, this covers the corner case of
a starting node being visited without having visited neighbours}. For parallel
coloring let us assume that there is a total ordering between colors, e.g.
color 1 comes before color 2, and so on.  A node keeps its color if it was
previously colored and becomes colored with the first unseen color in the
neighbourhood if it has the highest priority of all uncolored nodes in the
neighbourhood. For Kruskal's algorithm an edge remains in the MST it has been
already selected and an edge is selected to be inserted in the MST if all
lighter edges have been checked and the nodes, connected by that edge are not
in the same set. Note that for Kruskal's $\mathbf{x}^{(t)}$ ranges over the edges:

\begin{align}
    \text{BFS} : x_i^{(t+1)} &=
    \begin{cases}
        1 & x_i^{(t)}=1 \\
        1 & \exists j. (j, i)\in E \land x_j^{(t)}=1\\
        0 & \text{otherwise}
    \end{cases}
    \\
    \text{coloring} : x_i^{(t+1)} &= 
    \begin{cases}
        x_i^{(t)} & col_i^{(t)} \neq 0\\
        (col^{(t+1)}_i, p_i) & x_i^{(t)} = 0 \land p_i = \max\{p_j, j \in \mathcal{N}'_i \cup \{i\}\}\\ & \land \ col^{(t+1)}_i=\min\{col, col\notin seen_j\}\\
        (0, p_i) & \text{otherwise}
    \end{cases}
\end{align}
\begin{align}
    \text{Kruskal's} : x_i^{(t+1)} &= 
    \begin{cases}
        x_i^{(t)} & \text{if already selected}\\
        (1, w_i) & \text{if lighter edges visited} \land (a,b)=edge_i\\
                 & set^{(t)}(a) \neq set^{(t)}(b)\\
        (0, w_i) & \text{otherwise}
    \end{cases}
\end{align}
where for the coloring task $col_i^{(t)}$ denotes the coloring of node $i$ at
step $(t)$, $\mathcal{N}'_i$ denotes the set of uncolored neighbours, and
$seen_i$ denotes the set of colors observed in the neighbourhood of node $i$.
For Kruskal's, $(a, b)=edge_i$ denotes that edge $i$ connects nodes $a$ and
$b$, and $set^{(t)}(n)$ denotes the set of node $n$ at timestep $t$.
Before we proceed to the next step, if edge $edge_i=(a,b)$ is added to the MST,
we unify the sets of the two edge nodes, i.e. $set^{(t+1)}(a) = set^{(t+1)}(b)
= set^{(t)}(a) \cup set^{(t)}(b)$.

The ground-truth outputs we supervise towards are $\hat{y}_i^{(t)}=x_i^{(t+1)}$
for BFS, $\hat{y}_i^{(t)}=col_i^{(t+1)}$ for the coloring task and
$\hat{y}_i^{(t)}=x_i^{(t+1)}[0]$ for Kruskal's (i.e. 0/1 value whether each
edge is in the MST). We do not aim to reconstruct input node priorities or
input edge weights.


\section{Algorithms and concepts}\label{app:algosandconcepts}

\begin{table}
    \centering
    \caption{
        Algorithms and their corresponding concepts. Explanations are added to
        give a notion of how concepts can be used.
    }\label{tab:conceptsapp}
    \begin{tabular}{l l l}
        \toprule
        \multirow{2}{*}{\textbf{Algorithm}} & \multirow{2}{*}{\textbf{Concepts}} & \textbf{Example ground-truth explanations}\\
                                            &  & (not provided to the model)\\
        \midrule
        \multirow{2}{*}{BFS} & $hasBeenVisited$ ($hBV$) & $hVN(i)\implies y_i^{(t)}=1$\\
                             & $hasVisitedNeighbours$ ($hVN$) & $\exists i. \lnot hBV(i)\land hVN(i) \implies \tau^{(t)}=1$\\
        \midrule
        \multirow{4}{*}{coloring} &  & $iC(i)\land c1S(i) \land \lnot c2S(i) \implies y_i^{(t)}=2$ \\
                                  & \multirow{1}{*}{$isColored$ ($iC$), $hasPriority$ ($hP$)} &\\
                                  & \multirow{1}{*}{$colorXSeen$ ($cXS$), $X\in\{1,..,5\}$} &$(\lnot iC(i) \land hP(i)\land c1S(i) \land c2S(i)$\\
                                  & & $\land \ \lnot c3S(i)) \implies y_i^{(t)}=3$\\
        \midrule
        \multirow{5}{*}{Kruskal's} & & $(lEV(i)\land \lnot nISS(i) \land \lnot eIM(i))$ \\
                                  & $lighterEdgesVisited$ ($lEV$) & $ \implies y_i^{(t)}=1$ \\
                                  & \multirow{1}{*}{$nodesInSameSet$ ($nISS$)} &\\
                                  & \multirow{1}{*}{$edgeInMst$ ($eIM$)} & $(lEV(i) \land nISS(i) \land \lnot eIM(i))$\\
                                  & & $\implies y_i^{(t)}=0$\\
        \bottomrule
    \end{tabular}
\end{table}

Given the concepts algorithms in Table \ref{tab:conceptsapp}, here how each algorithm can be
executed with its own concepts:
\begin{itemize}
    \item BFS -- a node is visited if it has a visited neighbour and the
        execution continues until there is an unvisited node with visited
        neigbours.
    \item parallel coloring -- check if node has been colored, if not, check if
        it has priority to be colored on this step, if yes, check colors in the
        neighbourhood. Continue execution until a node with $\lnot isColored$
        exists.
    \item Kruskal's algorithm -- check if lighter edges have been visited by
        the algorithm, if not, edge is not in MST. Otherwise, check if the nodes
        connected by this edge belong in the same set and then if the edge has
        been previously selected. If the nodes are in the same set, the edge is
        in the MST for the next iteration if it has been selected for the MST
        previously. If nodes are not in the same set (this implies that the
        edge has not been selected for the MST), then this edge is in the MST
        from now on.
\end{itemize}

\section{Data generation}\label{app:data}
In general we aimed to maintain 10:1:1 train:validation:test split ratio and to
preserve the ratio across different data generation techniques.

For the BFS task we generate graphs from the following categories:
\begin{itemize}
    \item *Ladder graphs
    \item *2D grid graphs -- during generation we aimed the grid to be as close
        to square as possible
    \item Trees -- uniformly generated from the Pr\"{u}fer sequence \citep{Prufer1918Neuer}
    \item *Erd\H{o}s-R\'enyi \citep{Erdos1960Evolution}, $p$ of edge is $min\left(\frac{\log_2|V|}{|V|}, 0.5\right)$
    \item *Barab\'asi-Albert \citep{Albert2002Statistical} graphs, with either
        4 or 5 edges attached to each incoming node
    \item 4-Community graphs -- generated by creating 4 disjoint
        Erd\H{o}s-R\'enyi graphs with edge probability 0.7 and then
        interconnecting their nodes with probability of 0.01
    \item 4-Caveman \citep{Watts1999Networks} -- intra-clique edges are removed
        with $p=0.7$ and 0.025$|V|$ shortcut edges are inserted after that
        between cliques.
\end{itemize}
From each category we generate 100 graphs for training and 10 for
validation/testing. In total that is 700 for training and 70 for
validation/testing. To test strong generalisation, for a given number of nodes
in the graph, we generate 10 more test graphs of each category.

For the parallel coloring task we limit the number of colors to 5.
We generate 800/80/80 graphs for training/validation/testing with nodes of
fixed degree 5, so as to limit the possibility of 5 colors not being
enough. We pick the priority $p_i$ uniformly from the interval $[0, 255]$.

For Kruskal's, we generate 500 graphs of size 8 of each category with a star
from the list above for training and 50 for testing/validation of sizes $[8,
20, 50, 100]$.

To facilitate easier retention of self-information in the GNN, for the BFS and
parallel coloring tasks, we insert self-loops to all nodes for a given graph
similar to \citet{Velickovic2020Neural}. For Kruskal's, all nodes start with
a pointer to itself (see Appendix \ref{app:implementingkruskal})

\section{Implementing Kruskal's algorithm}\label{app:implementingkruskal}

\begin{figure}[t]
    \centering
    \includegraphics[width=\textwidth]{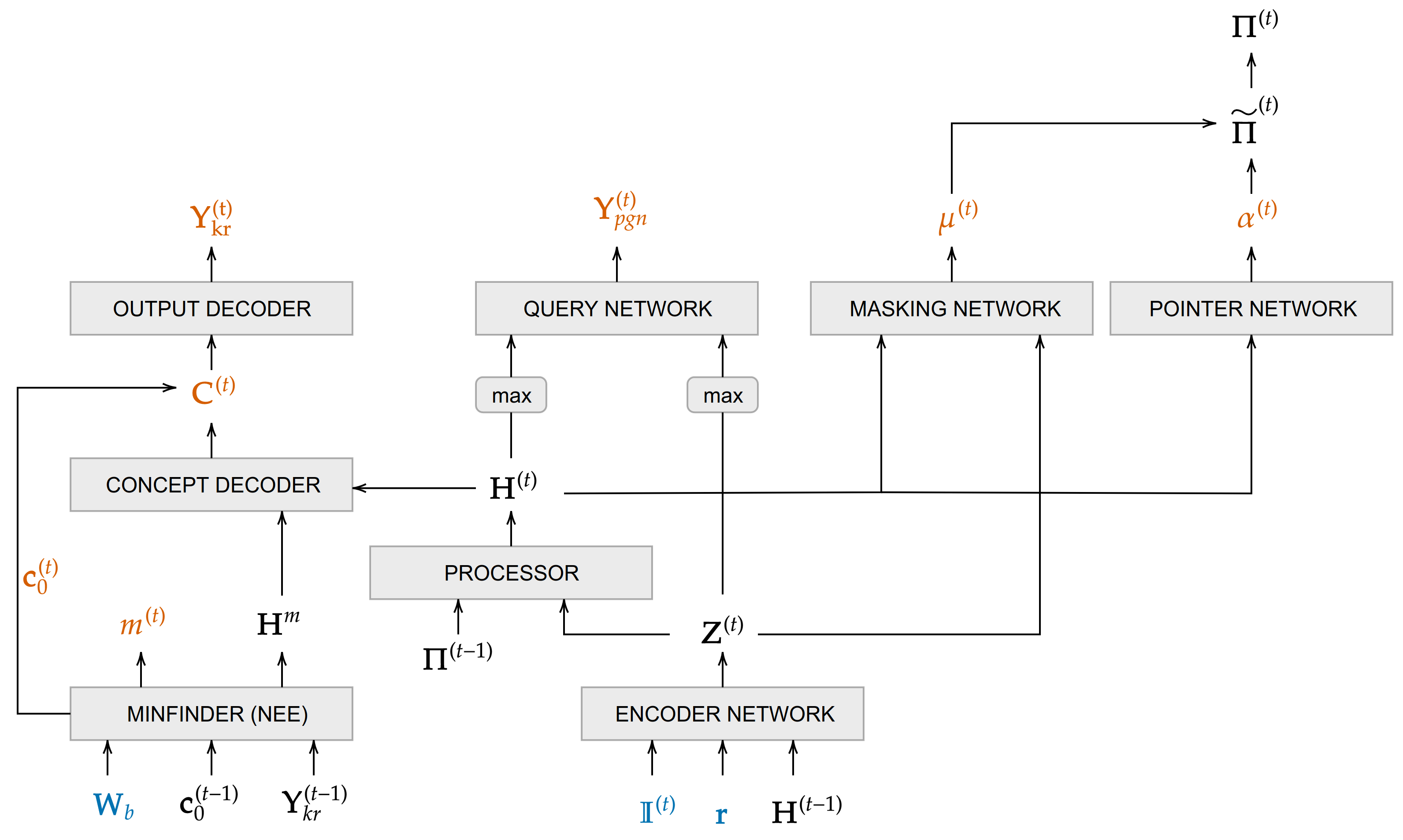}
    \caption{
        The dataflow for the Kruskal's algorithm. Blue variables are input
        variables, black inputs come from previous steps, supervision is
        performed on orange outputs.
    }\label{fig:kruskrus}
\end{figure}

For the implementation of the Kruskal's algorithm we need to implement
a min finding subroutine with masking and a disjoint-set union (DSU)
\citep{galler1964improved} data-structure.  We architecture the first as
a Neural Execution Engine (NEE) \citep{Yan2020Neural} that takes in the binary
edge weights $\mathbf{W}_b$, last step $lighterEdgesVisited$ concept for all
edges and last-step outputs $\mathbf{Y}^{(t-1)}_{kr}$. The NEE outputs
a pointer to the next minimal edge $m^{(t)}$ and the next $lighterEdgesVisited$
for all edges. We implement the DSU as a Pointer Graph Network (PGN)
\citep{Velickovic2020Pointer} that takes as input: (i) an indicator, which is
the two nodes of the minimal edge on the current step (teacher-forced during
training), (ii) node priorities (used as a tie-break when unifying sets) and
(iii) what's the last hidden state of the PGN algorithm $\textbf{H}^{(t-1)}$.
The PGN's processor also takes the last step pointers $\mathbf{\Pi}^{(t-1)}$ of
the DSU data structure. The PGN predicts: (i) whether the two nodes are in the
same set $\mathbf{Y}^{(t)}_{pgn}$; (ii) a mask $\mu^{(t)}$ over which nodes
should change their DSU pointers and (iii) an estimate of the new pointer
matrix.  The next step pointers are updated from (ii) and (iii) as follows:
\begin{equation}
    \mathbf{\widetilde{\Pi}}_{ij}^{(t)} = \mu_i^{(t)}\mathbf{\widetilde{\Pi}}_{ij}^{(t-1)} + \left(1-\mu_i^{(t)}\right)\mathbb{I}_{j=\text{argmax}_k\left(\alpha_{ik}^{(t)}\right)} \qquad \mathbf{\Pi}_{ij}^{(t)}=\mathbf{\widetilde{\Pi}}_{ij}^{(t)}\lor\mathbf{\widetilde{\Pi}}_{ji}^{(t)}
\end{equation}

To predict the rest of the concepts\footnote{apart from $lighterEdgesVisited$}
for \emph{every edge} $e=(u, v)$ we first concatenate the hidden state of the
first NEE transformer $\mathbf{H}^m_e$ with the hidden states of the PGN for
nodes $u$ and $v$, $\mathbf{H}^{(t)}_u$ and $\mathbf{H}^{(t)}_v$. The concept
decoder takes this as input and provides the rest of the concepts
$nodesInSameSet$ and $edgeInMst$, which is then passed to the output decoder to
produce final $\mathbf{Y}_{kr}^{(t)}$ whether each edge is selected in the MST
at step $t$.

The whole process is summarised in Figure \ref{fig:kruskrus}. Yellow variables
are those we supervise on. For a full list of losses used, see Appendix
\ref{app:algosandlosses}.



\section{Algorithms and respective losses}\label{app:algosandlosses}


We optimise our models based on the following losses:
\begin{itemize}
    \item Binary cross-entropy for concepts predictions
    \item Binary cross-entropy for predicting termination (when applicable to the algorithm)
\end{itemize}

Additionally for each algorithm, we add:
\begin{itemize}
    \item BFS:
        \begin{itemize}
            \item Binary cross-entropy for reachability predictions
        \end{itemize}
    \item parallel coloring: 
        \begin{itemize}
            \item categorical cross-entropy for reachability predictions
        \end{itemize}
    \item Kruskal's:
        \begin{itemize}
            \item categorical cross-entropy for predicting the position of the
                next selected edge
            \item binary cross-entropy for predicting whether two nodes are in
                the same set (cf. \citet{Velickovic2020Pointer})
            \item binary cross-entropy for predicting mask of the pointers to be
                changed (cf. \citet{Velickovic2020Pointer}) -- loss from positive examples is scaled by $\frac{1}{10}$
        \end{itemize}
\end{itemize}

\section{BFS accuracies}\label{app:BFSACC}

\begin{table}[t]
    \centering
    \caption{BFS accuracies over 5 runs}\label{tab:BFSACC}
    \begin{tabular}{l l r r r}
        \toprule
        \textbf{Model} & \textbf{Metric} & \multicolumn{1}{c}{$|V|=20$} & \multicolumn{1}{c}{$|V|=50$} & \multicolumn{1}{c}{$|V|=100$}\\
        \midrule

         \multirow{2}{*}{\shortstack[l]{Standard}}
            & mean-step acc. & $100.0{\scriptstyle\pm 0.0\%}$ & $100.0{\scriptstyle\pm 0.0\%}$ & $100.0{\scriptstyle\pm 0.0\%}$ \\
            & last-step acc. & $100.0{\scriptstyle\pm 0.0\%}$ & $100.0{\scriptstyle\pm 0.0\%}$ & $100.0{\scriptstyle\pm 0.0\%}$ \\
            & term. acc. & $100.0{\scriptstyle\pm 0.0\%}$ & $100.0{\scriptstyle\pm 0.0\%}$ & $100.0{\scriptstyle\pm 0.0\%}$ \\

                       \\
        \multirow{7}{*}{\shortstack[l]{Bottleneck\\(\small +next step pool)}}
                    & mean-step acc & $100.0{\scriptstyle\pm 0.0\%}$ & $100.0{\scriptstyle\pm 0.0\%}$ & $100.0{\scriptstyle\pm 0.0\%}$ \\
                    & last-step acc & $100.0{\scriptstyle\pm 0.0\%}$ & $100.0{\scriptstyle\pm 0.0\%}$ & $100.0{\scriptstyle\pm 0.0\%}$ \\
                    & term. acc. & $100.0{\scriptstyle\pm 0.0\%}$ & $100.0{\scriptstyle\pm 0.0\%}$ & $100.0{\scriptstyle\pm 0.0\%}$ \\
                    & formula mean-step acc. & $100.0{\scriptstyle\pm 0.0\%}$ & $100.0{\scriptstyle\pm 0.0\%}$ & $100.0{\scriptstyle\pm 0.0\%}$ \\
                    & formula last-step acc. & $100.0{\scriptstyle\pm 0.0\%}$ & $100.0{\scriptstyle\pm 0.0\%}$ & $100.0{\scriptstyle\pm 0.0\%}$ \\
                    & formula term. acc. & $100.0{\scriptstyle\pm 0.0\%}$ & $100.0{\scriptstyle\pm 0.0\%}$ & $100.0{\scriptstyle\pm 0.0\%}$ \\
                    & *concepts mean-step acc. & $100.0{\scriptstyle\pm 0.0\%}$ & $100.0{\scriptstyle\pm 0.0\%}$ & $100.0{\scriptstyle\pm 0.0\%}$ \\
                    & *concepts last-step acc. & $100.0{\scriptstyle\pm 0.0\%}$ & $100.0{\scriptstyle\pm 0.0\%}$ & $100.0{\scriptstyle\pm 0.0\%}$ \\

        \bottomrule
    \end{tabular}
\end{table}

We present BFS accuracy in Table \ref{tab:BFSACC}. Since BFS is relatively simple task
both the baseline and the concept-bottleneck model achieve perfect accuracy.

\section{Node-level decision tree}\label{app:DT-node}

\begin{figure}[t]
    \begin{subfigure}{.49\textwidth}
    \centering
    \includegraphics[width=1.7\textwidth, angle=0, scale=0.4]{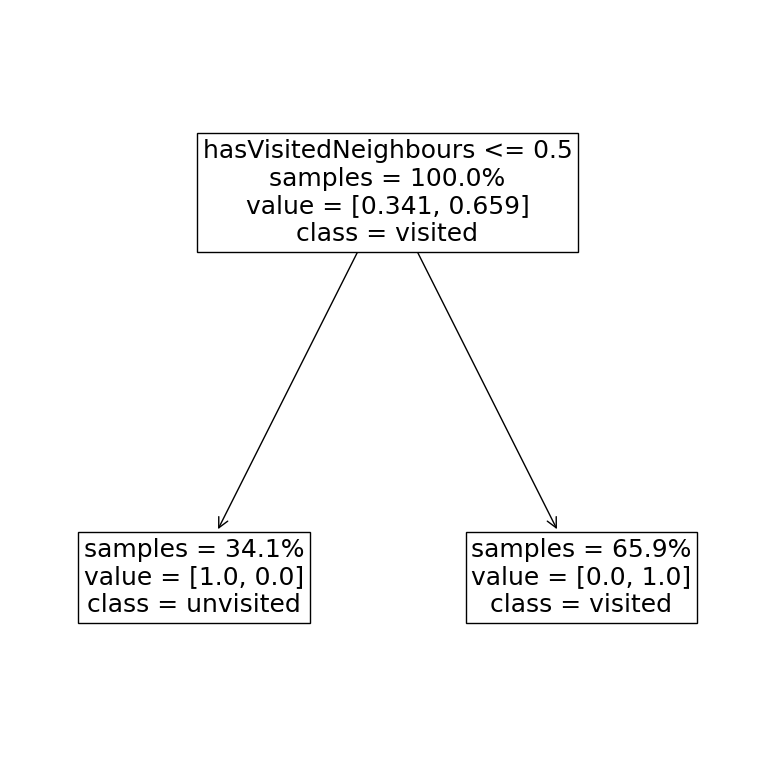}
    \caption{
        The decision tree for the BFS task. Due to the presence of self-loops
        the rule is simple -- check if visited neighbours exist (same node is
        always present in the neighbours list).
    }\label{fig:DT_BFS}
    \end{subfigure}
    \begin{subfigure}{.49\textwidth}
    \centering
    \includegraphics[width=1.7\textwidth, angle=0, scale=0.4]{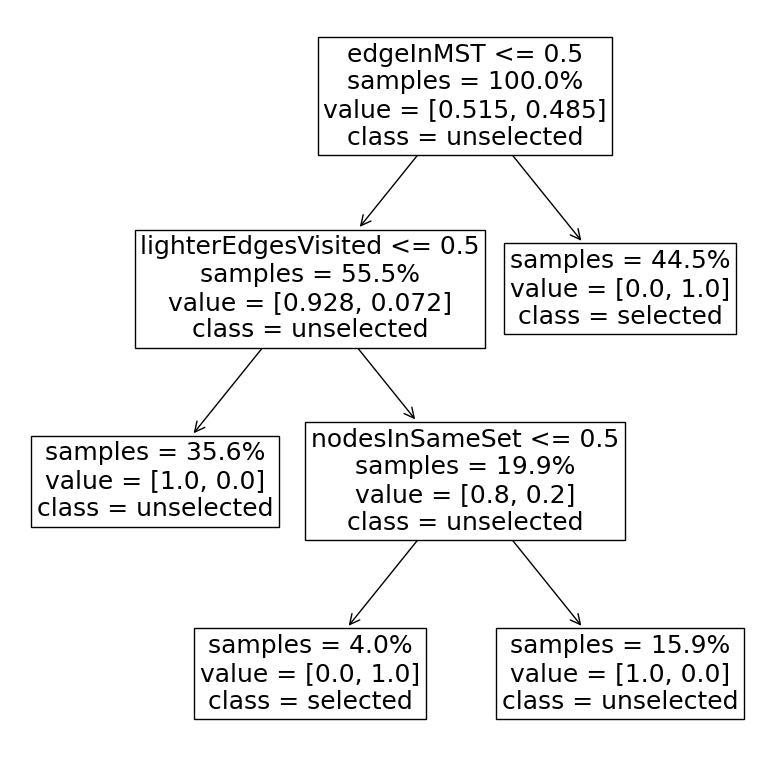}
    \caption{
        The decision tree for the Kruskal's algorithm.
    }\label{fig:DT_kruskal}
    \end{subfigure}
    \caption{Decision trees for BFS and Kruskal's}
\end{figure}

\begin{figure}[t]
    \centering
    \includegraphics[width=0.9\textwidth, angle=90, scale=1.70]{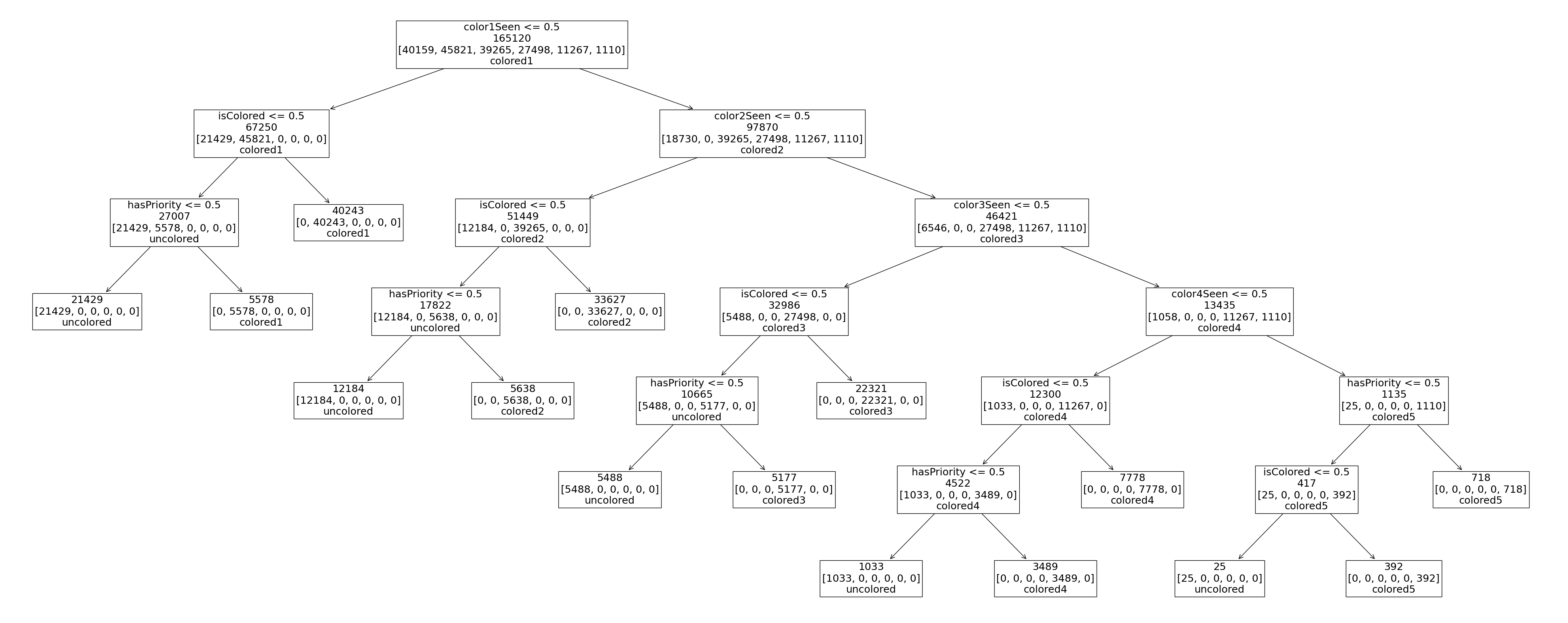}
    \caption{
        The decision tree for the parallel coloring task. The classifier first
        checks colors seen around, then priority and only then, whether a node
        is colored or not. 
    }\label{fig:DT_PC}
\end{figure}

The decision trees are shown in Figures \ref{fig:DT_BFS}\&\ref{fig:DT_PC}. In
all cases the rules follow the logic of the original algorithm. It should
be noted that the decision tree for the BFS task consists of just a single
concept -- why are concepts necessary then? The answer is simple -- the 
addition of termination requires us to use the concepts $hasBeenVisited$.

Although not displayed in Figure \ref{fig:DT_PC} due to space constraints, the
decision tree helped discover a 'bug' -- one of the concepts ($color5Seen$) was
never used, which suggests that it may not be necessary.

\section{Oracle model for parallel coloring}\label{app:oraclecoloring}

\begin{figure}[t]

    \begin{subfigure}{0.48\textwidth}
        \includegraphics[width=\textwidth]{./figures/concept_supervised_coloring_predinet_L1AndPruned_per_concept_mean_acc.png}
        \caption{Concept mean-step accuracy}
    \end{subfigure}
    \begin{subfigure}{0.48\textwidth}
        \includegraphics[width=\textwidth]{./figures/concept_supervised_coloring_predinet_L1AndPruned_per_concept_last_acc.png}
        \caption{Concept last-step accuracy}
    \end{subfigure}

    \begin{subfigure}{0.48\textwidth}
        \includegraphics[width=\textwidth]{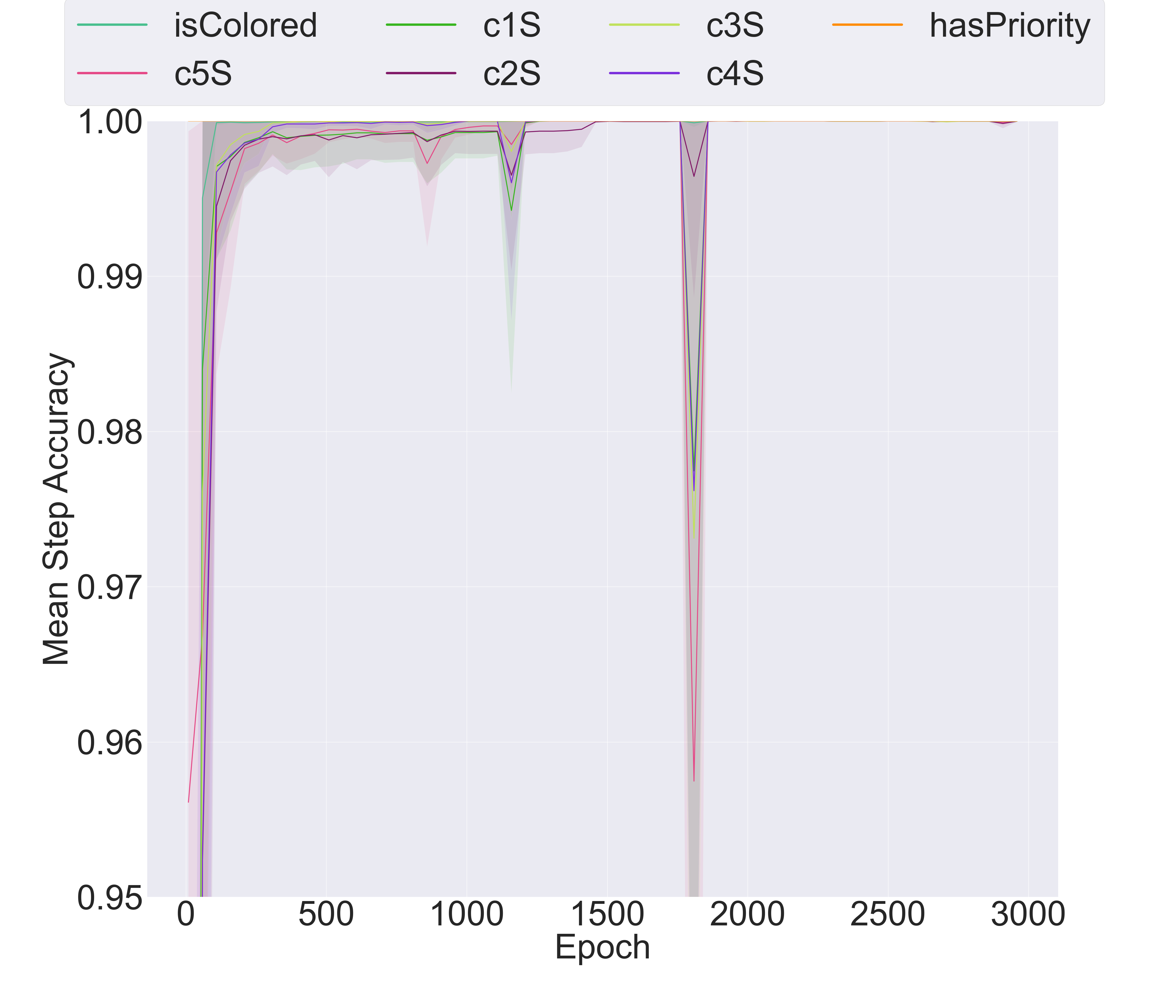}
        \caption{Concept mean-step accuracy, $hasPriority$ provided}
    \end{subfigure}
    \begin{subfigure}{0.48\textwidth}
        \includegraphics[width=\textwidth]{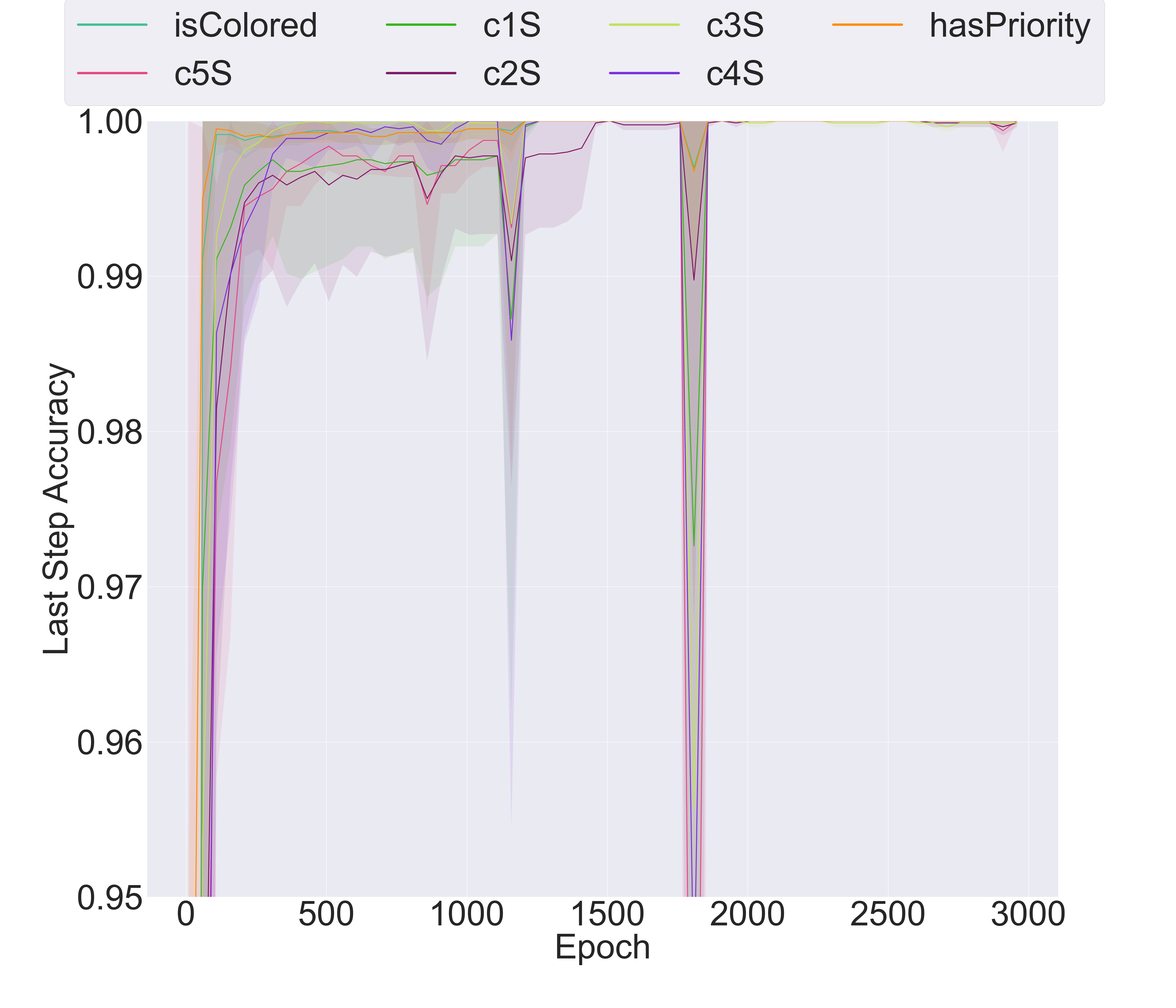}
        \caption{Concept last-step accuracy, $hasPriority$ provided}
    \end{subfigure}
    \caption{
        Concept accuracies per epoch of the parallel coloring algorithm with
        and without hardcoding the $hasPriority$ concept. Clearly, hardcoding
        this concept makes our morel much more stable.
    }\label{fig:concept_step_acc_hardcoded1}
\end{figure}
In the main body of the paper (Figure \ref{fig:concept_step_acc}) we showed
that quite often there were drops of accuracy for almost all concepts for the
parallel coloring task. Based on more careful observation it was hypothesised
that the $hasPriority$ is \emph{too critical} for the task. To prove this we
plot the per-step concept accuracy of an oracle model that is always provided
with the correct value for that concept. Figure
\ref{fig:concept_step_acc_hardcoded1} clearly shows the importance of the
$hasPriority$ concept -- hardcoding its value to the ground-truth allows us to
learn the task almost perfectly. 

\section{Example explanations of algorithms}\label{app:algo-explanations}

\begin{table}[t]
    \centering
    \caption{
        BFS explanations. The model explanations match the ground-truth ones.
        Available concepts are $hasVisitedNeighbours$ and $hasBeenVisited$.
        Max number of explanation occurrences is 5.
    }\label{tab:BFSEXPL}
     
    \begin{tabular}{l l r}
        \toprule
        \textbf{Thing to explain} & \textbf{Explanation} & \multicolumn{1}{c}{\textbf{\# of occurrences}}\\
        \midrule
        $n$ is NOT visited & {\scriptsize $\lnot hasBeenVisited(n) \land \lnot hasVisitedNeighbours(n)$} & 5 \\
        $n$ is visited & {\scriptsize $hasVisitedNeighbours(n)$} & 5 \\
        \\
        continue execution & {\scriptsize $\lnot hasBeenVisited(n) \land hasVisitedNeighbours(n)$} & 5 \\

        \bottomrule
    \end{tabular}
\end{table}
\begin{table}[t]
    \centering
    \caption{
        Kruskal's explanations. The model explanations match the ground-truth
        ones. Available concepts are $nodesInSameSet$, $lighterEdgesVisited$ ($lEV$)
        and $edgeInMst$. Max number of explanation occurrences is 5.
    }\label{tab:kruskrusEXPL}
     
    \begin{tabular}{l l r}
        \toprule
        \textbf{Thing to explain} & \textbf{Explanation} & \multicolumn{1}{c}{\textbf{\# of occurrences}}\\
        \midrule
        \multirow{2}{*}{$e$ NOT in MST} & {\scriptsize $(nodesInSameSet(e) \land \lnot edgeInMst(e))\lor$} & \multirow{2}{*}{5} \\
                                        & {\scriptsize $(\lnot lEV(e)\land \lnot edgeInMst(e))$} & \\
                                        \\
        \multirow{4}{*}{$e$ in MST} & {\scriptsize $(lEV(e) \land edgeInMst(e))\lor$} & \multirow{2}{*}{1} \\
                                    & {\scriptsize $(lEV(e)\land \lnot nodesInSameSet(e))$} & \\
                                    \\
                                    & {\scriptsize $(lEV(e) \land nodesInSameSet(e) \land edgeInMst(e))\lor$} & \multirow{2}{*}{4} \\
                                    & {\scriptsize $(lEV(e)\land \lnot nodesInSameSet(e) \land \lnot edgeInMst(e))$} & \\
        \bottomrule
    \end{tabular}
\end{table}

\begin{table}[t]
    \centering
    \caption{
        Bottleneck model without pruning/L1 loss gives applicable accuracies --
        formula accuracies are as high as their counterparts. But are they as
        interpretable?
    }\label{fig:PCNOREG}
    \begin{tabular}{l l r r r}

        \toprule
        & & \multicolumn{3}{c}{\textbf{Parallel coloring}}\\
          & & \multicolumn{1}{c}{$|V|=20$} & \multicolumn{1}{c}{$|V|=50$} & \multicolumn{1}{c}{$|V|=100$}\\
        \midrule
            & mean-step acc. & $99.55{\scriptstyle\pm 0.17\%}$ & $99.39{\scriptstyle\pm 0.22\%}$ & $99.22{\scriptstyle\pm 0.32\%}$ \\
            & last-step acc. & $99.33{\scriptstyle\pm 0.23\%}$ & $99.06{\scriptstyle\pm 0.37\%}$ & $98.95{\scriptstyle\pm 0.52\%}$ \\
            & term. acc. & $99.69{\scriptstyle\pm 0.27\%}$ & $99.41{\scriptstyle\pm 0.21\%}$ & $99.07{\scriptstyle\pm 0.38\%}$ \\
            & formula mean-step acc. & $99.55{\scriptstyle\pm 0.18\%}$ & $99.39{\scriptstyle\pm 0.22\%}$ & $99.22{\scriptstyle\pm 0.33\%}$ \\
            & formula last-step acc. & $99.31{\scriptstyle\pm 0.18\%}$ & $99.03{\scriptstyle\pm 0.36\%}$ & $98.93{\scriptstyle\pm 0.52\%}$ \\
            & formula term. acc. & $99.66{\scriptstyle\pm 0.27\%}$ & $99.41{\scriptstyle\pm 0.21\%}$ & $99.07{\scriptstyle\pm 0.38\%}$ \\
            & concepts mean-step acc. & $99.81{\scriptstyle\pm 0.08\%}$ & $99.68{\scriptstyle\pm 0.11\%}$ & $99.6{\scriptstyle\pm 0.16\%}$ \\
            & concepts last-step acc. & $99.61{\scriptstyle\pm 0.16\%}$ & $99.38{\scriptstyle\pm 0.2\%}$ & $99.27{\scriptstyle\pm 0.27\%}$ \\
         \bottomrule
    \end{tabular}
\end{table}

\begin{table}[t]
    \centering
    \caption{
        Some example parallel coloring explanations for node $n$, \emph{no
        regularisation applied}. (For brevity we omit listing every single
        explanation.) Class explanations do not match the ground truth and
        sometimes contain usage of variables, whose value can be inferred from
        the values of other variables. Available concepts are $isColored$
        ($iC$), $hasPriority$ ($hP$) and $colorXSeen, X\in\{1,..,5\}$. Max
        number of explanation occurrences is 5.
    }\label{tab:PCEXPL}
     
    \begin{tabular}{l l r}
        \toprule
        \textbf{Thing to explain} & \textbf{Explanation} & \multicolumn{1}{c}{\textbf{\# of occurrences}}\\
        \midrule
        \multirow{3}{*}{$n$ is not colored} & {\scriptsize $(\lnot iC(n) \land \lnot hP(n) \land \lnot color5Seen(n)) \lor$} & \multirow{3}{*}{1} \\
                                            & {\scriptsize $(color5Seen(n) \land color4Seen(n) \land \lnot iC(n) \land \lnot hP(n) $} &  \\
                                            & {\scriptsize $\lnot color1Seen(n) \land \lnot color2Seen(n))$} &  \\
        \\
        \multirow{2}{*}{$n$ has color 2} & {\scriptsize $(iC(n) \land \lnot hP(n) \land color1Seen(n) \land \lnot color2Seen(n)) \lor$} & \multirow{2}{*}{2}\\
                                         & {\scriptsize $(hP(n) \land color1Seen(n) \land \lnot iC(n) \land \lnot color2Seen(n))$} &  \\
        \\
        \multirow{4}{*}{$n$ has color 5} & {\scriptsize $(hP(n) \land color2Seen(n) \land color3Seen(n)$} & \multirow{4}{*}{2}\\
                                         & {\scriptsize $ \land color4Seen(n) \land \lnot color5Seen(n) \land \lnot iC(n)) \lor$} &  \\
                                         & {\scriptsize $(iC(n) \land color1Seen(n) \land color2Seen(n) \land $} &  \\
                                         & {\scriptsize $color3Seen(n) \land color4Seen(n) \land \lnot color5Seen(n))$} &  \\
        \\
        \multirow{4}{*}{$n$ has color 5} & {\scriptsize $(hP(n) \textcolor{red}{\land color1Seen} \land color2Seen(n) \land color3Seen(n)$} & \multirow{4}{*}{3}\\
                                         & {\scriptsize $ \land color4Seen(n) \land \lnot color5Seen(n) \land \lnot iC(n)) \lor$} &  \\
                                         & {\scriptsize $(iC(n) \land color1Seen(n) \land color2Seen(n) \land $} &  \\
                                         & {\scriptsize $color3Seen(n) \land color4Seen(n) \land \lnot color5Seen(n))$} &  \\

        \bottomrule
    \end{tabular}
\end{table}

\begin{table}[t]
    \centering
    \caption{
        Parallel coloring explanations for node $n$, \emph{with regularisation
        applied}. Class explanations are concise, consistent across seeds and
        very close to the ground truth. Available concepts are $isColored$
        ($iC$), $hasPriority$ ($hP$) and $colorXSeen, X\in\{1,..,5\}$. Max
        number of explanation occurrences is 5.
    }\label{tab:PCCORRECTEXPL}
     
    \begin{tabular}{l l r}
        \toprule
        \textbf{Thing to explain} & \textbf{Explanation} & \multicolumn{1}{c}{\textbf{\# of occurrences}}\\
        \midrule
        \multirow{1}{*}{$n$ is not colored} & {\scriptsize $\lnot iC(n) \land \lnot hP(n) $} & \multirow{1}{*}{5} \\
        \\
        \multirow{2}{*}{$n$ has color 1} & {\scriptsize $(iC(n) \land \lnot hP(n) \land \lnot color1Seen(n)) \lor$} & \multirow{2}{*}{5}\\
                                         & {\scriptsize $(hP(n) \land \lnot color1Seen(n) \land \lnot iC(n))$} &  \\
        \\
        \multirow{7}{*}{$n$ has color 2} & {\scriptsize $(iC(n) \land \lnot hP(n) \land color1Seen(n) \land \lnot color2Seen(n)) \lor$} & \multirow{2}{*}{4}\\
                                         & {\scriptsize $(hP(n) \land color1Seen(n) \land \lnot color2Seen(n)\land \lnot iC(n))$} &  \\
         \\
                                         & {\scriptsize $(iC(n) \land \lnot hP(n) \land color1Seen(n) \land \lnot color2Seen(n)) \lor$} & \multirow{4}{*}{1}\\
                                         & {\scriptsize $(hP(n) \land color1Seen(n) \land \lnot iC(n)\land \lnot color2Seen(n)) \lor$} &  \\
                                         & {\scriptsize $(hP(n) \land color4Seen(n) \land \lnot iC(n)\land$} &  \\
                                         & {\scriptsize $\lnot color2Seen(n)\land \lnot color3Seen(n))$} &  \\
        \\
        \multirow{4}{*}{$n$ has color 3} & {\scriptsize $(iC(n) \land \lnot hP(n) \land color1Seen(n)$} & \multirow{4}{*}{5}\\
                                         & {\scriptsize $\land color2Seen(n) \land \lnot color3Seen(n)) \lor$} &\\
                                         & {\scriptsize $(hP(n) \land color1Seen(n)\land color2Seen(n)$} &\\
                                         & {\scriptsize $\land \lnot color3Seen(n) \land \lnot iC(n))$} & \\
        \\
        \multirow{4}{*}{$n$ has color 4} & {\scriptsize $(iC(n) \land \lnot hP(n) \land color1Seen(n)$} & \multirow{4}{*}{5}\\
                                         & {\scriptsize $\land color2Seen(n) \land color3Seen(n) \land \lnot color4Seen(n)) \lor$} &\\
                                         & {\scriptsize $(hP(n) \land color1Seen(n) \land color2Seen(n)$} &\\
                                         & {\scriptsize $\land color3Seen(n) \land \lnot color4Seen(n) \land \lnot iC(n))$} & \\
        \\
        \multirow{4}{*}{$n$ has color 5} & {\scriptsize $(iC(n) \land \lnot hP(n) \land color1Seen(n)$} & \multirow{4}{*}{5}\\
                                         & {\scriptsize $\land color2Seen(n) \land color3Seen(n) \land color4Seen(n)) \lor$} &\\
                                         & {\scriptsize $(hP(n) \land color1Seen(n) \land color2Seen(n)$} &\\
                                         & {\scriptsize $\land color3Seen(n) \land color4Seen(n) \land \lnot iC(n))$} & \\

        \bottomrule
    \end{tabular}
\end{table}

We list example explanations for BFS in Table \ref{tab:BFSEXPL} and for Kruskal's. Our
observations are that for simpler algorithms with fewer possible concept
combinations, such as BFS, explanations match \emph{very closely} the ground
truth -- the only difference is that unvisited rule relies on $\lnot
hasBeenVisited$. This, however, would not break the execution -- in fact
the network learns to use this concept to reinforce its outputs since
$\lnot hasBeenVisited$ always holds when $\lnot hasVisitedNeighbours$.

For the parallel coloring task, we first draw the attention to the results
without using any regularisation techniques in Table \ref{fig:PCNOREG}. The
main observation is that regularisation is not necessary to achieve high
accuracy -- formula-based metrics do not deviate largely from formulas obtained
without using metrics. However, as can be seen from Table
\ref{tab:PCEXPL} when the possible combinations and
outputs grow, some combinations may not be observed during training and
therefore the explanations may contain unnecessary concepts (e.g., whether
color 5 is seen in the exlanation for uncolored node $n$) or sometimes not
consider some concept (red in Table \ref{tab:PCEXPL}). Given the applications
of the formulas still achieved very high accuracy, our hypothesis was that such
algorithm require regularisation techniques, such as adding auxiliary L1 loss
and pruning, in order to force the model to rely on as few concepts as
possible. Results confirmed our hypothesis -- Table \ref{tab:PCCORRECTEXPL}
shows that 4/5 times \emph{all} explanations are correct and only once
the rule for color 2 contained an extra clause.

\section{Software} \label{app:software}
The code for the experiments is implemented in Python 3, relying upon open-source libraries such as Scikit-learn (BSD license) \citep{pedregosa2011scikit} and Pytorch (BSD license) \citep{paszke2019pytorch}. All the experiments have been run on an NVIDIA Titan Xp 12 GB GPU.



\end{document}